\documentclass[final,1p,times]{elsarticle}

\usepackage{amssymb}
\usepackage{multirow}
\usepackage{siunitx}
\usepackage{subcaption}
\usepackage{amsmath}
\usepackage{url}

\journal{Acta Astronautica}

\begin{document}

\begin{frontmatter}



\title{A Simulation-Augmented Benchmarking Framework for Automatic RSO Streak Detection in Single-Frame Space Images}

%

\author[1]{Zhe Chen}
\ead{zhe.chen1@sydney.edu.au}
\author[2,3]{Yang Yang}	
\ead{yiyinfeixiong@gmail.com}
\author[3]{Anne Bettens}
\ead{anne.bettens@sydney.edu.au}
\author[3]{Youngho Eun}
\ead{youngho.eun@sydney.edu.au}
\author[3]{Xiaofeng Wu$^*$}
\cortext[cor1]{Corresponding author}
\ead{xiaofeng.wu@sydney.edu.au}
\fntext[1]{School of Computer Science, the University of Sydney, Australia}
\fntext[2]{School of Mechanical and Manufacturing Engineering, University of New South Wales, Australia}
\fntext[3]{School of Aerospace, Mechanical and Mechatronic Engineering, the University of Sydney, Australia}

%
            




\begin{abstract}
Effective space-based Resident Space Objects (RSOs) detection provides a solution to avoiding potential impact or collision of space debris with other satellites. This is achieved by providing timely observations that help track and infer the behaviours of targets of interest. Recently, deep convolutional neural networks (DCNNs) have shown superior performance in object detection compared to traditional methods when large-scale datasets are available to train their models. However, collecting rich, diversified, and meaningful data of RSOs is difficult due to very few occurrences in the space images. Without sufficient data, it is challenging to comprehensively train DCNN detectors and make them effective for detecting RSOs in space images, let alone to evaluate and estimate whether a detector has sufficient capacity to detect RSOs robustly. 
The lack of meaningful evaluation of different detectors could further affect the design and application of detection methods.
To tackle this issue, we propose that the space images containing RSOs can be simulated to complement the shortage of raw data for better benchmarking. Accordingly, we introduce a novel simulation-augmented benchmarking framework for RSO detection (SAB-RSOD). In our framework, by making the best use of the hardware parameters of the sensor that captures real-world space images, we first develop a high-fidelity RSO simulator that can generate various realistic space images to enrich the dataset for training detectors. Then, we use this simulator to generate images that contain diversified RSOs in space and annotate them automatically. Later, we mix the synthetic images with the available real-world images, obtaining around 500 images for training detectors with only the real-world images for evaluation. Under SAB-RSOD, we can train different popular object detectors like  Yolo\cite{redmon2016you} and Faster RCNN\cite{ren2015faster} effectively, enabling us to evaluate their performance thoroughly. The evaluation results have shown that the amount of available data and image resolution are two important factors for robust RSO detection. Moreover, if using a lower resolution for higher efficiency, we demonstrated that a simple  UNet-based\cite{ronneberger2015u} detection method can already access high-accuracy RSO detection, providing new insights to the community. The dataset and evaluation code will be released.
\end{abstract}




\end{frontmatter}



\section{Introduction}\label{1}

It has become a concern in recent years that the near-Earth orbits are turning into a congested and contaminated environment.
With the advent of satellite mega-constellations, the number of active and operational satellites surges to over 8,000 by the first quarter of 2022 according to the Space Environment Statistics conducted by European Space Agency (ESA) Space Debris Office\,\cite{esa2022esa}. More concerning is the proliferation of orbital debris. Unfortunately, space surveillance networks and maintained catalogs do not include most space debris objects. Any impact or collision of these debris objects with other operational satellites can cause severe damage or even end their life. This may trigger the so-called Kessler Syndrome, which creates more additional collisions and can cause catastrophic problems for space missions. For instance, the well-known space collision between the inactive Russian communication satellite Cosmos 2251 and the active commercial communication satellite Iridium 33 in 2009 produced almost 2,000 pieces of debris objects, which poses continual and critical threats to other active Resident Space Objects (RSOs). This kind of catastrophic event is becoming increasingly more likely as the number of public and private satellites in orbit increases \cite{marty2023space}.

To avoid such a space traffic problem in orbit, there is an increasing need to observe all satellites, particularly space debris, determine their positions, and update their orbital parameters with high accuracy \cite{vasudhara2022initial}. The number of sensors, primarily optical, but also radar and radio frequency both ground and space-based being used for tracking RSOs in the Space Situational Awareness (SSA) context has been growing. 
Usually, optical-based space tracking consists of two scenarios: the staring (or survey) mode and the chasing (or tracking) mode. In the former tracking scenarios, the optical sensor is pointing towards the sky while moving at the sidereal rate or in orbit. In particular, the space-based sensor has a nominal pointing direction with a pre-launch attitude profile unless it is controlled to rotate during the mission. Since the angular velocity of the space objects, especially those in the lower Earth orbit is larger than the background stars, they appear as streaks of light in the Field of View (FOV) while stars appear as dots on the optical image. 

For SSA applications, space-based optical systems have advantages over ground-based optical data collection. For example, challenges with time-of-day lighting are somewhat mitigated, and weather/atmospheric conditions are hardly an issue. Optical sensors in space are also more sensitive and allow for detecting dimmer objects including space debris. Several space missions have demonstrated the use of such sensors for tracking RSOs. The Midcourse Space Experiment satellite was launched by the US Air Force in 2000 with the Space-Based Visible sensor, which was working as a Space Surveillance Network (SSN) node to track satellites and debris in and close to geosynchronous orbit\,\cite{sharma2002toward}. Canada's NEOSSat satellite was equipped with the world's first space telescope to monitor near-Earth objects including RSOs, which successfully demonstrates the detection of self-conjunction RSOs\,\cite{abbasi2019neossat}. Besides, a lighthouse concept has also been proposed by ESA to design a telescope with a diameter of 50 cm to be installed on a small satellite platform for space surveillance and tracking purposes\,\cite{lupo2018lighthouse}. Furthermore, most satellites are equipped with at least one optical sensor, such as a star tracker, for attitude determination purposes. There is interest in investigating the feasibility of applying the star tracker for SSA purposes by leveraging its dual-use mode. For example, Spiller et al.\,\cite{spiller2020onorbit} studied the star sensor together with a dedicated algorithm for recognizing RSOs onboard, which has been tested via synthetic images provided by a high-fidelity simulator. 

Once optical images are taken, the detection of targets of interest, i.e., RSOs on the images, becomes an essential task. Generally, RSO detection can be categorized into single-frame and multi-frame methods. The single-frame methods leverage filtering techniques\,\cite{levesque2009automatic, vananti2020streak} or morphology\,\cite{kim2016astride,virtanen2016streak,nir2018optimal} to extract targets from images. While multi-frame methods rely on inter-frame information, normally \textit{a priori} trajectory information of space objects across a set of consecutive images to judge the existence of the target. A faint streak detection method based on a streak-like spatial filter was proposed in Ref.\,\cite{vananti2020streak}. However, its time consumption of 2h to 3h for processing a single $4000\times4000$ pixels image is relatively unaffordable. 
In a single image, some researchers formulate streak detection as a line extraction task, but this heavily relies on the assumption that the streak is of significant length and has clear appearance for detection. For example, streaks can be detected using line extraction techniques like Hough or Radon transforms\cite{ginkel2004hough}. 
Ref.\,\cite{kim2016astride} introduced the Automated Streak Detection for Astronomical Images (ASTRiDE) algorithm for astronomical images based on image segmentation methods. Despite some promising results, these methods usually have a high failure rate in handling RSO images with a low signal-to-noise ratio (SNR). The StreakDet algorithm presented in Ref.\,\cite{virtanen2016streak} employs a Hough-like optimized line-extraction algorithm to locate streak features, but the work does not characterize the detected streak by providing centroid and length information. Ref.\,\cite{nir2018optimal} shows the Radon transform provides a good solution to extract streaks in a signal-maximizing way. However, Radon transform mostly presents its effectiveness for longer exposure time images, such as astronomical images, where the streaks are spreading over 50 pixels. Ref.\,\cite{spiller2020onorbit} introduced a star sensor image processing framework for space object detection by using traditional methods. Despite the aforementioned advancements in RSO detection, challenges are still presented if the RSO streaks are insufficiently long or their SNRS are high.

Until recent years, Deep Convolutional Neural Network (DCNN) based object detectors are attracting more interest than traditional detectors, which can be used to predict centroid and length of streaks effectively. On generic object datasets\,\cite{lin2014microsoft,everingham2010pascal} which include objects like people, cars, and apples, DCNN-based detectors like Faster RCNN (Faster Regional Convolutional Neural Network)\,\cite{ren2015faster}, FPN (Feature Pyramid Network)\,\cite{lin2017feature}, and Yolo (You-only-look-once)\,\cite{redmon2016you} have shown impressive performance in areas, such as accuracy and efficiency, outperforming traditional object detectors that rely on pre-defined operations like Hough or Radon transforms to a great margin. In the RSO detection task, one of the major goals is to study the feasibility of different DCNN-based object detection algorithms for detecting RSOs that are less like generic objects. There are a few papers that successfully applied generic object detectors to detect space-related objects. In a related paper\,\cite{jia2020detection}, Faster RCNN with a modified ResNet-50 backbone was utilized to detect astronomical targets, which achieved 80.8\% average precision. Another related study aims to identify supernovae using FCOS(Fully Convolutional One-Stage Object Detection)\,\cite{yin2021supernovae}, which achieved 99.4\% and 71.3\% F1 scores on two different datasets: PSP-SN and PSP-Ori, respectively. Some other related studies introduce another segmentation task to detect objects at pixel-level. For example, researchers used the UNet\,\cite{ronneberger2015u,zhou2019unet++} which generates DCNN outputs with the same resolution of the input image to detect small objects like point sources and spheroids \cite{hausen2020morpheus}. Researchers also used Mask RCNN\,\cite{he2017mask} which can perform instance segmentation to identify galaxy and star objects\,\cite{burke2019deblending}, by which 49.0\% average precision was achieved. 
Notwithstanding this progress, we argue that applying these detectors on RSOs, which are generally dim and have small streaks in space images, has not been thoroughly explored. Moreover, since available data for RSO detection is very limited, and the aforementioned DCNN-based detectors are generally data-hungry, studying the effects of different detectors on RSO detection is quite challenging. In particular, we mainly found that the Swarm star tracker images\,\cite{jorgensen2017swarm} which has the data for RSO detection, can only collect several dozens of valid images from this source. Without rich data, we found that popular detectors that are effective for identifying generic objects generally do not perform very well for RSO detection with limited data for training, making it difficult to understand whether a detector is more feasible for this task. It is also worth mentioning that, although the use of DCNNs can be effective in determining the relative orbit of an RSO, most of this work is done while the spacecraft is already in orbit and close proximity is required to capture images of the RSOs \cite{gong2022deep, guthrie2022image, koenig2021artms}, and comprehensive and detailed sensor information is often required \cite{oakes2022double, guthrie2022image} to apply DCNN techniques for SSA purposes effectively.

Using only the images provided by the Swarm star tracker, one significant challenge facing machine learning is the scarcity of large, reliable datasets. This is especially true in the domain of space, making the task of RSO detection more difficult \cite{song2022deep}. Instead of gathering and annotating real-world data, which can be time-consuming and labour-intensive, we suggest creating a \emph{simulation-augmented framework} for benchmarking and evaluating detectors on the RSO detection task. More specifically, based on the Swarm images, simulating an environment that has a virtual sensor inside with the identical parameters of the real sensor used to collect Swarm data. In this simulated environment, we employ plenty of virtual RSOs in space which appear as diversified streaks in the space images obtained from this simulated environment. Such simulated images can also annotate the appeared streaks automatically. Then, we mix these simulated images with real-world images together for training different detectors. The trained detectors are later used to detect RSOs from the remaining real-world images for evaluation. We call such benchmarking procedure a simulation-augmented benchmarking framework for RSO detection (SAB-RSOD). 

To sum up, the main contributions of this work are summarised below:
\begin{itemize}
   \item Based on the sensor parameters from the real-world, a simulation system is developed to simulate space-based images capturing both stars and RSOs, where RSOs appear as streaks. This simulation can also annotate streaks automatically.
   \item A new simulation-augmented space-based space tracking dataset and a benchmark are proposed for evaluating different RSO detectors. To our best knowledge, this is the first time that real-world Swarm star tracker (SST) images augmented by synthetic images have been used to train and validate DCNN-based detectors on RSO detection. 
   \item Two popular detectors like Faster RCNN\,\cite{ren2015faster} and Yolo\,\cite{redmon2016you} are evaluated. Additionally, a simple UNet-based object detector is further developed for a better comparison study.
   \item With simulated data, we can better reveal the effects of evaluated detectors on the RSO detection task. Under the SAB-RSOD, we can validate that the amount of data for training and image resolutions that are important in generic object detection continues to be the two core factors that have a great impact on performance. In addition, we found that a simple UNet-based detector performs surprisingly well on low-resolution images despite a much smaller capacity than compared popular detectors. We believe that these findings can deliver new insights to the researchers in this community. 
\end{itemize}


\section{Benchmarking Framework}\label{sec:statePro}

\subsection{Overview}
To augment the SSA capability by leveraging space-based optical tracking sensors, there is a need to develop algorithms and software that can automatically detect and report the presence of RSO streaks in the acquired images. Different from ground-based optical tracking, whose datasets are easier to obtain, it is difficult to collect space-based images, particularly for those including both stars and RSOs. As a result, it is pivotal to develop a well-labelled dataset as well as a framework that can leverage state-of-the-art DCNNs for automatic RSO streak detection.
\begin{figure*}[t!]
\centering
\includegraphics[width=0.95\textwidth]{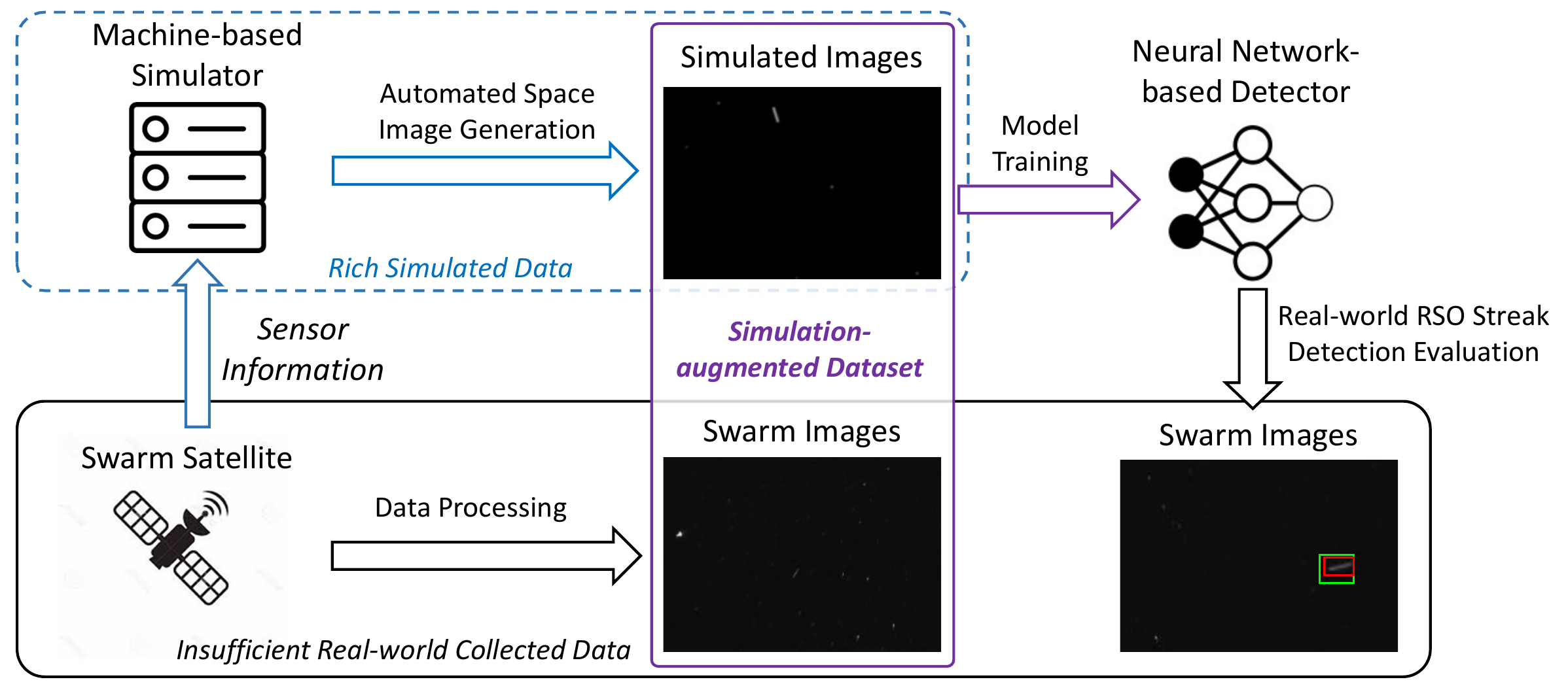}
\caption{The proposed simulation-augmented benchmarking framework for RSO detection  (SAB-RSOD). Based on the Swarm platform \cite{jorgensen2017swarm} that provides real-world data for RSO detection, we use the Swarm satellite sensor information to develop a machine-based simulator. We then apply our simulator to generate space images that contain automatically annotated RSOs. By mixing the generated space images with the real-world images, we can then perform effective training for different DCNN-based detectors. Later, we evaluate the performance of the trained DCNN-based detectors on real-world data to help reveal the capacities of different DCNN-based detectors regarding the RSO detection task.}
\label{fig:satDetFramework}
\end{figure*}

In this study, we developed a simulation-augmented benchmarking framework to tackle the issues mentioned above. The detailed framework processing can be found in Fig.\,\ref{fig:satDetFramework}. In general, our benchmarking procedure is mainly based on the  real-world raw images provided by the Swarm dataset. We process the data by filtering out images that do not contain any RSOs and annotating the appeared RSOs in the filtered images, obtaining a new dataset that is ready for training and testing an RSO detection algorithm. Meanwhile, based on the sensor parameters that are identical to the sensor used to capture Swarm data, we develop a high-fidelity image simulator to generate synthetic images. With these synthetic images, we perform training on different detection methods. Lastly, the trained detection methods are all evaluated on the test set of the real-world Swarm images. We follow the typical 5-fold cross-validation to perform train and testing split on the real-world Swarm images. 

In the rest of this paper, we will subsequently introduce the Swarm data, the image simulator, and the DCNN-based detection methods used for comparison under our SAB-RSOD framework. Experimental results are presented in the following section. 

\subsection{RSO Streak Detection and Swarm Data}

Firstly, we formulate the RSO streak detection in space images as follows: an object detector predicts the categories, locations, and sizes of RSO streaks appearing in an image captured in the space. Hence, it involves both classification and regression. For RSO detection particularly, there is only one category of object that is of interest. Therefore, the desired output for a space object detector is a 4-dimensional descriptor: $(x,y,w,h)$, where $(x,y)$ means the horizontal and vertical coordinate of the top left corner of a space object, or a streak, in the image, and $(w,h)$ means its width and height, respectively. 

\begin{figure*}[!ht]
\centering
\includegraphics[width=0.8\textwidth]{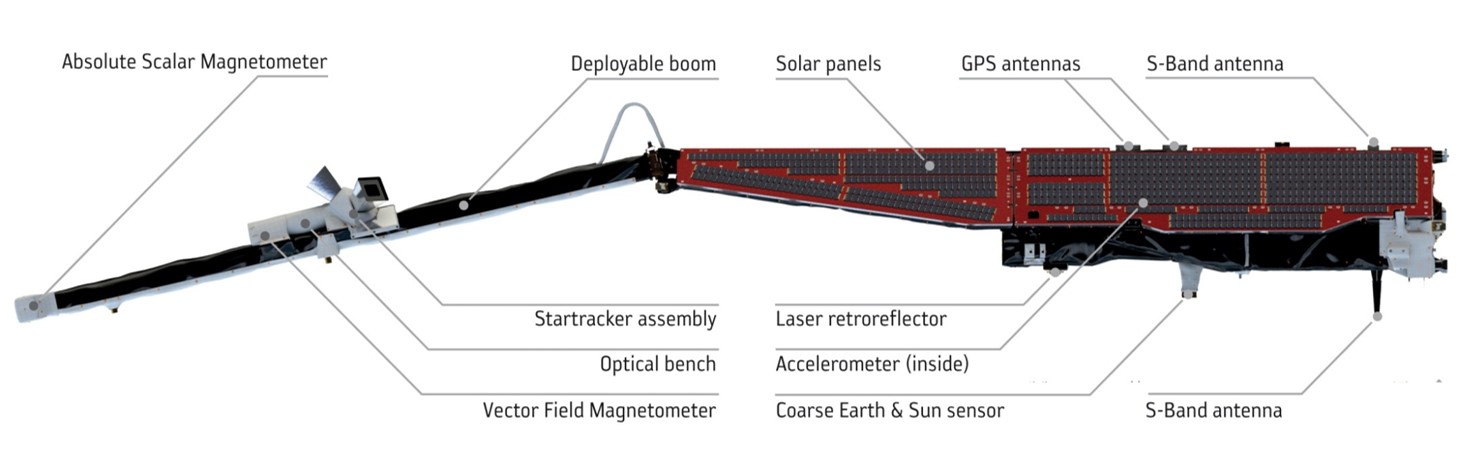}
\caption{The Swarm satellite and its instruments.}
\label{fig:swarm_satellite}
\end{figure*}

However, as mentioned previously, the publicly available space images that contain meaningful RSOs to detect are too limited. 
To the authors' best knowledge, only a few data sources of space-based tracking images are useful and accessible for RSO streak detection. 
Thanks to the Swarm satellite constellation operated by ESA for earth observation, images taken from the onboard star trackers are retrieved and can be collected to build a proper dataset. The mission consists of three identical satellites (A, B, and C). A list of payloads is installed onboard (see Fig.\,\ref{fig:swarm_satellite}), among which the $\mu\text{ASC}$ star tracker is the focus of this paper\,\cite{jorgensen2017swarm}. The star tracker is comprised of three Camera Head Units (CHUs) mounted on the innermost end of the optical bench. 
Some key parameter of the $\mu\text{ASC}$ star tracker are given in Table\,\ref{tab:sstPara}\,\cite{eisenman1996real}.

Images taken by CHUs are downlinked to the ground and Feiteirinha et al. processed them to explore the opportunity of RSO tracking\,\cite{feiteirinha2019str4sd} and published on GitHub\,\footnote{\url{https://github.com/esa/str4sd}}. Among them, we collected a total of 63 images with at least one clear streak which was then annotated manually. One example for such an image is shown in Fig.\,\ref{fig:exSwarmImg}. More detail on pre-processing the dataset is presented in Appendix\,\ref{sec:appendix}. Extracted images from this real-world dataset are used for testing and validating the proposed benchmarking framework in this work.

\begin{table}[ht]
    \caption{Key parameters of Swarm $\mu\text{ASC}$ star tracker.}
    \centering
    \begin{tabular}{l|r}
        \hline
        Effective focal length        & \SI{19980}{\mu\meter} \\
        \hline
        Number of pixels in x ($n_x$) & 752 \\
        \hline
        Number of pixels in y ($n_x$) & 580 \\
        \hline
        Pixel length in x ($x_p$)     & \SI{8.6}{\mu\meter} \\
        \hline
        Pixel length in y ($y_p$)     & \SI{8.3}{\mu\meter} \\
        \hline
        Integration time              & \SI{1.0}{\second}\\
        \hline
    \end{tabular}
    \label{tab:sstPara}
\end{table}

Although the Swarm dataset is valuable and useful for testing RSO detection, it is still a small amount. It is reported that current popular and effective object detection methods\,\cite{ren2015faster, redmon2016you,redmon2018yolov3} generally require a large amount of data\,\cite{deng2009imagenet, lin2014microsoft} to optimise their model parameters and ensure promising performance. Without diversified data, it poses challenges to identify what is important for designing an effective deep learning method to detect RSO streaks. In particular, it is a task that needs sufficient dim, small, and rotated streaks. To tackle this issue, we further develop an image simulator that can automatically generate space objects, or streaks, together with stars in a space-based optical image. 

In summary, to accomplish the task of RSO streak detection from each individual image, a simulation-aided Swarm data benchmarking framework is proposed, which leverages multiple state-of-the-art DCNN methods.

\begin{figure}[!ht]
    \centering
    \includegraphics[width=0.4\textwidth]{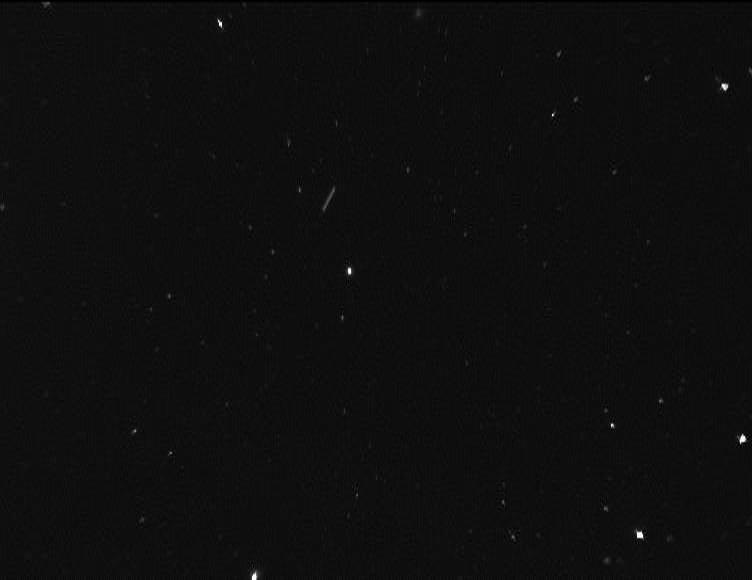}
    \caption{One Exemplar Image from Swarm Dataset}
    \label{fig:exSwarmImg}
\end{figure} 
\section{Swarm Image Simulation with RSO Streaks}\label{sec:simulator}

\begin{figure*}[!ht]
\centering
\includegraphics[width=0.65\textwidth]{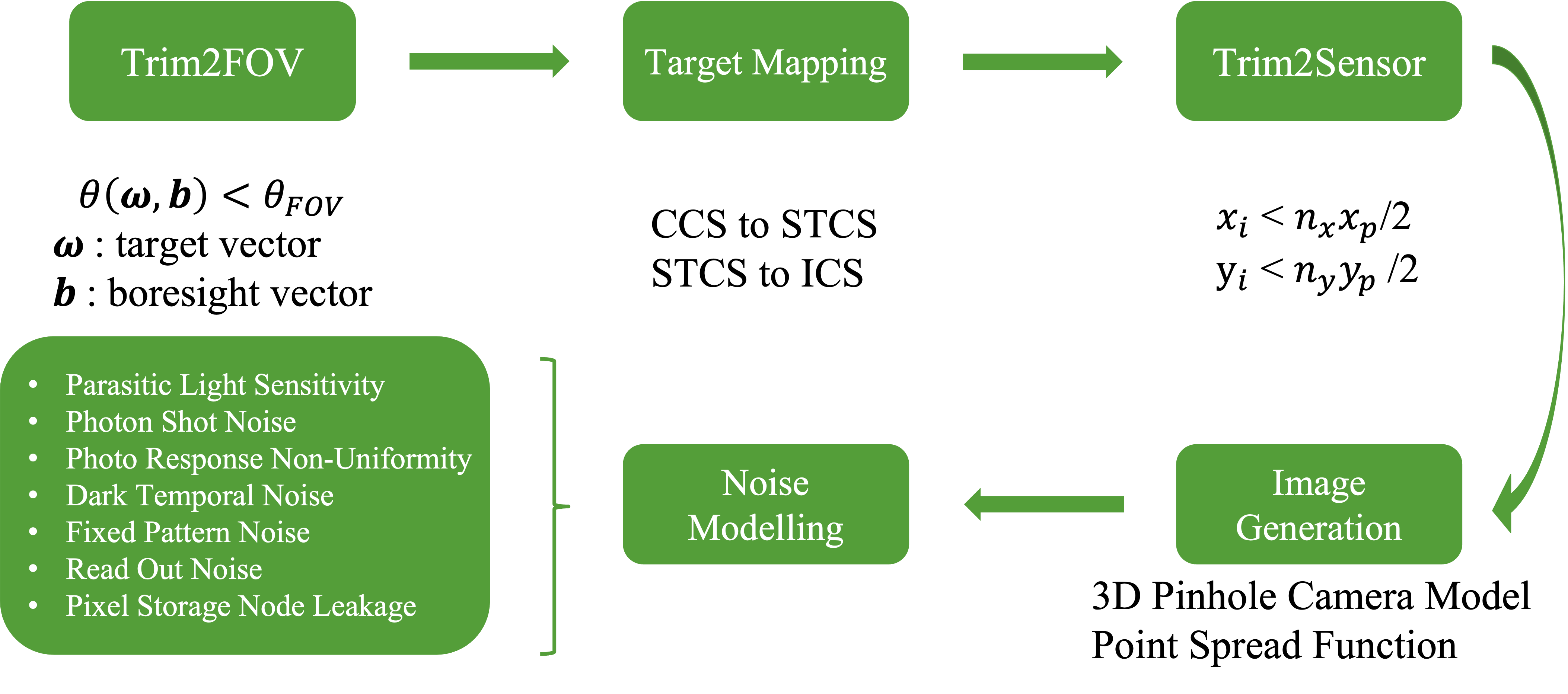}
\caption{\label{fig:img_sim} Overall chart for space image simulation.}
\end{figure*}

To provide the training data for effective RSO streak detection, an image simulator is developed to simulate space images with stars and RSOs by electro-optical (EO) sensors, such as telescopes and star trackers. It is capable of simulating two types of tracking scenarios, i.e., ground-based and space-based in general. Space-based EO sensors can be placed in low Earth orbit (LEO) to track RSOs nearby. 
The general mathematical models used in the simulator are similar to those presented in Refs.\,\cite{smith2017development} for star tracker simulation. In addition, the streak model for RSO has also been implemented in the simulator.
Fig.\,\ref{fig:img_sim} presents the skeleton diagram for the whole image simulation process, which consists of five major steps. The first step is to determine if the object (a star or a space object) will fit into the FOV of the lens. Afterwards, the target, with its coordinate represented in the inertial celestial coordinate system (astrometric position), will be converted into the coordinates of the sensor frame. If the object meets the sensor frame constraints, an image will be generated based on two principles: the three-dimensional pinhole camera model and the point spread function. Finally, various types of noises are added to the uncontaminated image, forming a realistic and reliable image.

\subsection{Measuring Principle of Star Tracker}
\begin{figure}
    \centering
    \includegraphics[width=0.3\textwidth]{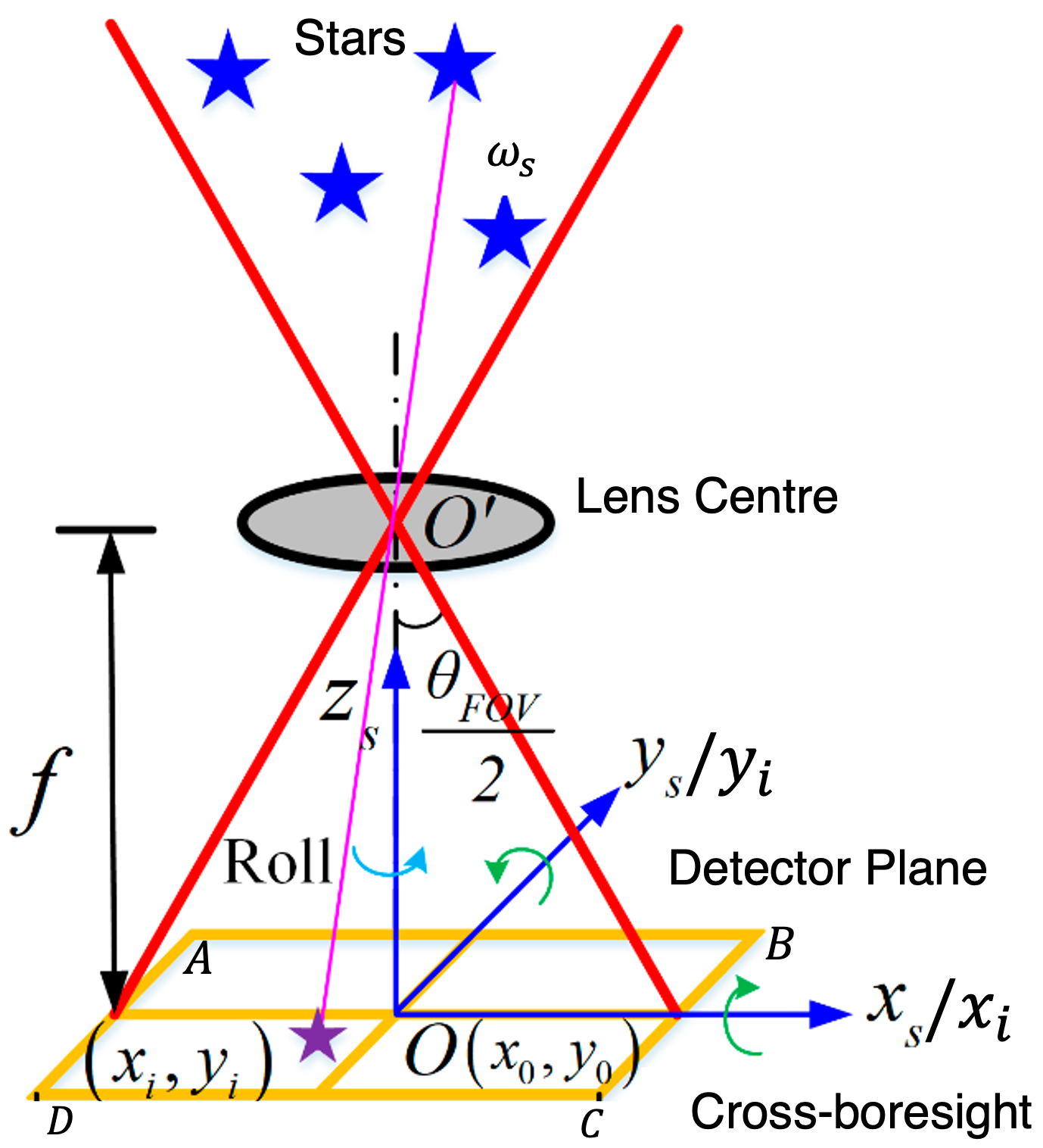}
    \caption{Illustration of the star tracker coordinate system, the image coordinate system, and front projection imaging\,\cite{liu2021compressed}.}
    \label{fig:sensorFrame}
\end{figure}

Three coordinate systems are used: the celestial coordinate system (CCS) uses right ascension ($\alpha$) and declination ($\delta$) to represent the location of a star. The star tracker coordinate system (STCS) is located at the centre (principal point $\mathcal{O}(x_0, y_0)$) of the detector plane indicated by the quadrangle $\mathcal{ABCD}$ with its $z_s$ axis aligning with the boresight and $x_s$ and $y_s$ axes parallel to both sides of the detector plane. The distance between the detector centre and the lens centre $\mathcal{O}'$ is defined as the focal length $f$. The image coordinate system (ICS) $(x_i, y_i)$ is a 2-dimensional frame coinciding with the STCS on the detector plane. See Fig.\,\ref{fig:sensorFrame} for more details about the measuring principle of the star tracker. 
Given a star coordinate $(\alpha, \delta)$ obtained from existing star catalogues and the attitude of the star tracker represented by three Euler angles: right ascension $(\alpha_0$, declination $\delta_0$ and roll $\phi_0)$ known \textit{a priori}. The corresponding image coordinate $(x_i, y_i)$ can be calculated by two steps: transforming $(\alpha_i, \delta_i)$ to the star tracker coordinate $(x_s, y_s)$ and then conducting the perspective projection. 
If the star appears in the star tracker FOV indicated by the quadrangle $\mathcal{ABCD}$, i.e., the following conditions are met, 
\begin{equation}
    \theta < \frac{\theta_{\mathcal{FOV},x}}{2}\,\cap\,\theta < \frac{\theta_{\mathcal{FOV},y}}{2},
\end{equation}
where $\theta$ is the angular distance between the boresight vector $\Vec{b}$ and the target vector $\Vec{\omega}$, both in the CCS. The boresight vector is calculated from the star tracker Euler angles:
\begin{equation}
    \Vec{b} = \begin{bmatrix}
        \cos{\delta_0}\cos{\alpha_0}\\
        \cos{\delta_0}\sin{\alpha_0}\\
        \sin{\delta_0}
    \end{bmatrix}.
\end{equation}
Similarly, $\Vec{\omega}$ is calculated from the astrometric position of the target $(\alpha, \beta)$.
The length/width FOVs are calculated via:
\begin{align}
    \theta_{\mathcal{FOV},x} = 2 \arctan\Big(\frac{n_{x} x_p}{2f}\Big),\\
    \theta_{\mathcal{FOV},y} = 2 \arctan\Big(\frac{n_{y} y_p}{2f}\Big),
\end{align}
where $n_{x}$ and $x_p$ indicate the number of pixels and the length of each pixel in $x$ axis while $n_{y}$ and $y_p$ indicate those values in $y$ axis.

The target vector in the STCS is calculated by:
\begin{equation}
    \Vec{\omega}_s = 
    \begin{bmatrix}
        x_s\\
        y_s\\
        z_s
    \end{bmatrix} = 
    M\begin{bmatrix}
        \cos{\delta}\cos{\alpha}\\
        \cos{\delta}\sin{\alpha}\\
        \sin{\delta}
    \end{bmatrix},
\end{equation}
where $M$ is the rotation matrix from the CCS to the STCS.
According to the pinhole camera model, the mapping from $\Vec{\omega}_s$ to the image plane is given by:
\begin{equation}
    s\begin{bmatrix}
        x_i\\
        y_i\\
        1
    \end{bmatrix} = 
    \begin{bmatrix}
        f & 0 & x_0\\
        0 & f & y_0\\
        0 & 0 & 1
    \end{bmatrix}
    \begin{bmatrix}
        x_s\\
        y_s\\
        z_s
    \end{bmatrix},
    \label{eq:imgCoor}
\end{equation}
where s is a scale factor. The calculated image coordinates $x_i, y_i$ have to meet with:
\begin{equation}
    x_i < n_{x} x_p\,\cap\,y_i < n_{y} y_p,
\end{equation}
in order to appear on the image plane. 

\subsection{Simulation of Targets}
The simulation of star and RSO accounts for the target magnitude, the optics and the geometric characteristics of the detector. 
\subsubsection{Star Catalogue}
Star's positions, normally represented by right ascension and declination angles, can be obtained from existing star catalogues, such as the Yale Bright Star Catalogue\,\cite{bahcall1987analysis}. This catalogue lists a total of 9,095 stars with stellar magnitude 6.5 or brighter, which is roughly every star visible to the naked eye from Earth. Another larger catalogue Hipparacos is established by extracting scientific data taken from the European Space Agency's Hipparcos satellite, with over 118,000 stars and a limiting magnitude of 12.4\,\cite{perryman1997hipparcos}. Its improved version, Hipparcos-2 catalogue, achieves accuracy for nearly all stars brighter than a magnitude of 8, which is up to a 4 factor improvement than is precursor \,\cite{leeuwen2007validation}. Both of these star catalogues have been implemented in this work. Considering the star magnitude limit is given as 6.5 in the simulation, the difference between sky maps generated by two catalogues can be neglected. 
\subsubsection{RSO Catalogue}
Different from star catalogues, existing RSO catalogues tend to record the RSO's orbit in the format of orbital elements or Cartesian coordinates. Among them, the so-called two-line element (TLE) format is a \textit{de facto} standard for the distribution of RSO orbital elements in near-Earth space. TLEs and radar cross-section areas for most RSOs can be retrieved from SpaceTrack\,\footnote{https://www.space-track.org} or Celestrack\,\footnote{https://celestrak.org} websites and they are used to calculate the astrometric positions and visual/apparent magnitudes based on the cannonball assumption.
\subsubsection{Image Pixel Generation}
To simulate star tracker images, the brightness or magnitudes of stars in the FOV must be converted to a corresponding number of electrons ($n_e$). Normally, a point spread function (PSF) describes the response of an imaging system to a point source (e.g., a star), whose Gaussian image is smaller than the central maxima of the diffraction pattern. 
\paragraph{Point Spread Function}
The uncorrelated Gaussian PSF is expressed as: 
\begin{equation}
    f(x,y)=\frac{A}{2 \pi \sigma_x \sigma_y }\exp \left(-\left({\frac {(x-x_i)^{2}}{2\sigma_{x}^{2}}}+{\frac {(y-y_i)^{2}}{2\sigma _{x}^{2}}}\right)\right), 
    \label{eq:gaussianPSF}
\end{equation}
where $A$ is the amplitude, $x_i$ and $y_i$ are the pixel locations obtained from Eq.\ref{eq:imgCoor}, and $\sigma_x$ and $\sigma_y$ are the corresponding Gaussian variances. The integral of Eq.\,\ref{eq:gaussianPSF} leads to the number of electrons:
\begin{equation}
    n_e = 2\pi A \sigma_x\sigma_y.
    \label{eq:eNumber}
\end{equation}
Hence the amplitude $A$ can be calculated as:
\begin{equation}
    A = \frac{n_e}{2\pi  \sigma_x\sigma_y}.
    \label{eq:eNumber2}
\end{equation}
Afterwards, the Gaussian PSF in Eq.\ref{eq:gaussianPSF} can be used to randomly generate responding pixels in the ICS for a specific star. 
\paragraph{Calculation of Electron Numbers}
It is well known the star magnitude $m_*$ can be formulated via the irradiance $I_*$ compared to those values of the Sun:
\begin{equation}
    m_{*}=m_{\text{Sun}}-2.5\log _{10}\left({\frac {I_{*}}{I_{\text{Sun}}}}\right),
\end{equation}
where $m_{\text{sun}}$ is the Sun's apparent magnitude observed from the Earth and $m_{\text{sun}} = -26.7$.

\begin{figure}[!ht]
    \centering
    \includegraphics[width=0.45\textwidth]{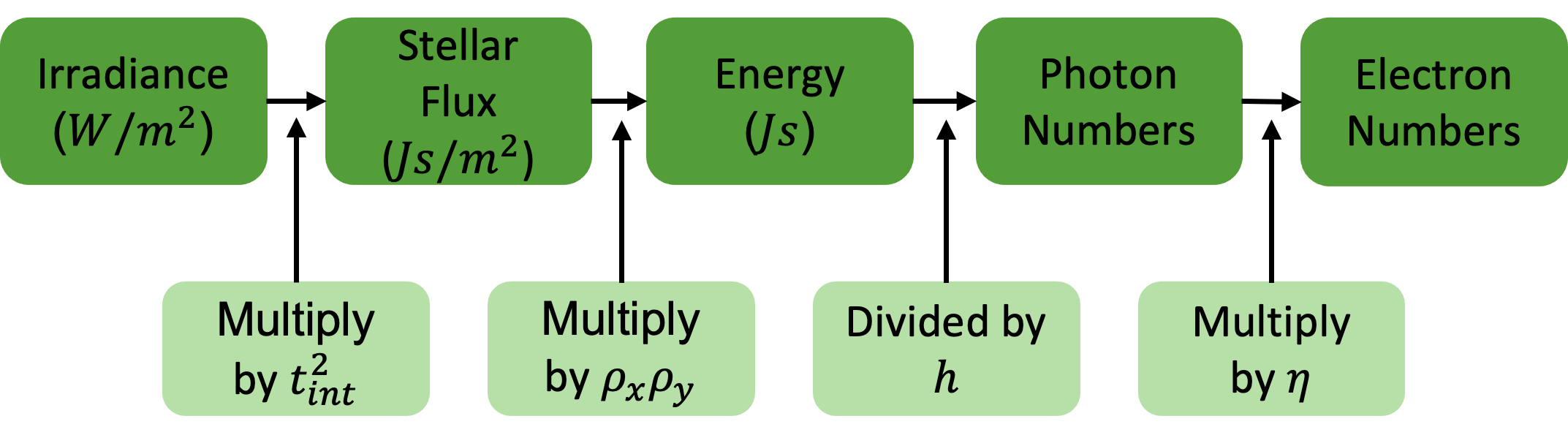}
    \caption{Conversion process from flux density to the associated number of electrons\,\cite{smith2017development}.}
    \label{fig:flux2eNumber}
\end{figure}
\begin{figure*}[!th]
    \centering
    \includegraphics[height=0.3\textheight, width=\textwidth]{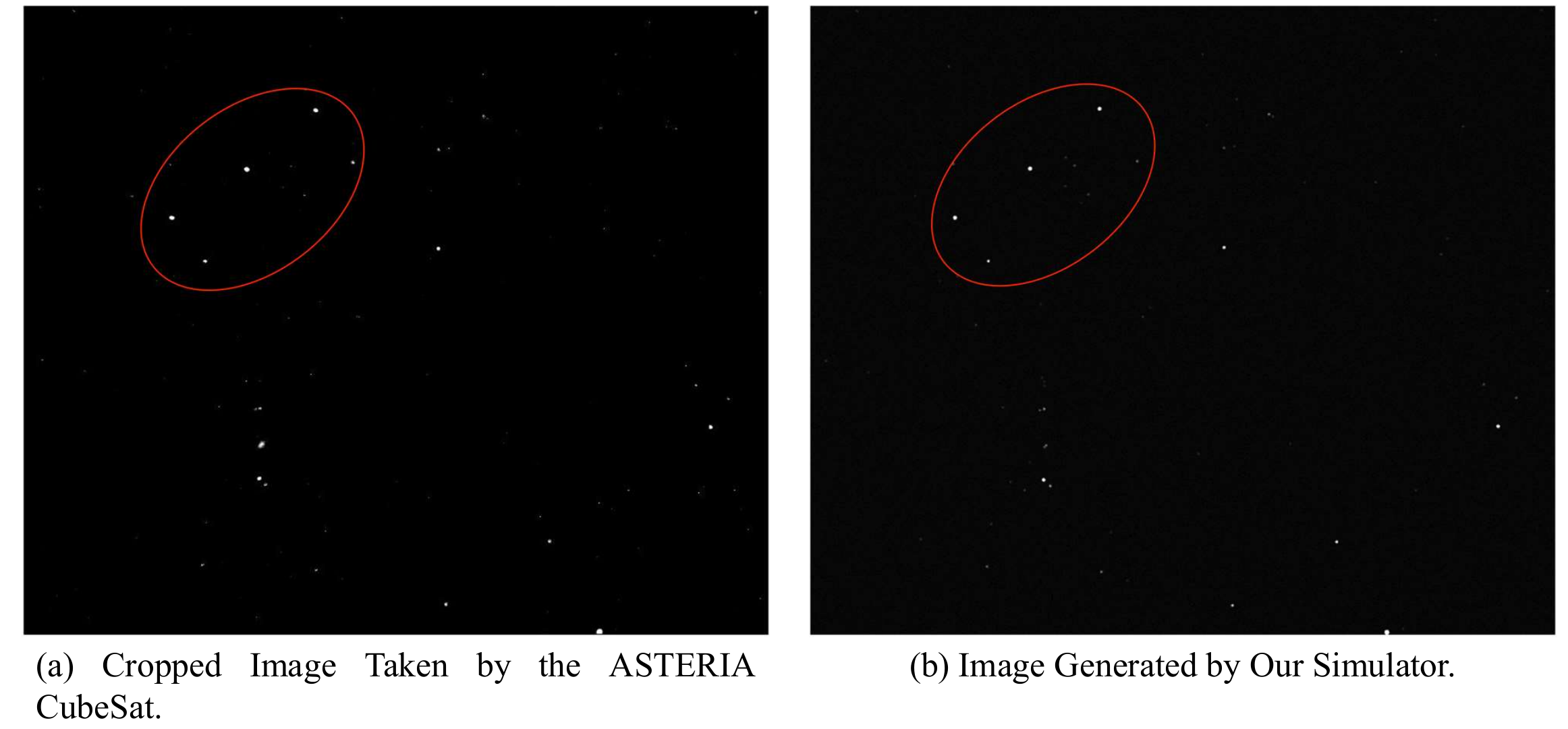}
    \caption{Validation via ASTERIA CubeSat image.}
    \label{fig:ussisvalidation}
\end{figure*}

More specifically, the visual magnitude of the space object is calculated by\,\cite{hejduk2011specular}:
\begin{gather}
    m_{\text{RSO}} = m_{\text{sun}} + 5log(\frac{r\Delta}{a^2}) - 2.5log(\beta F_{\text{diff}}(\phi) \notag\\
    + (1-\beta) F_{\text{spec}}(\phi)).
\end{gather}
The visual magnitude of the sun is denoted by $m_{\text{sun}}$, $p$ is the reflectivity of the RSO (technically geometric albedo), $R$ is the radius of the RSO in km, $a$ is 1 AU in km, $r$ is the distance from the RSO to the Sun in km, $\Delta$ is the distance from the RSO to the star tracker in km, $\phi$ is the solar phase angle (the angle between the sun and the star tracker relative to the RSO), $\beta$ is the diffusion coefficient and $F_{\text{diff}}$ and $F_{\text{spec}}$ are the diffusive and specular components of the solar phase angle function, respectively.
Some combinations of the specular and diffuse phase function components are given as below\,\cite{hejduk2011specular}: 
\begin{gather}
    F_{\text{spec}}(\phi) = \frac{1}{4},\\
    F_{\text{diff}}(\phi) = \frac{2}{3\pi}\big[(\pi-\phi)cos(\phi)+sin(\phi)\big].
\end{gather}
Finally, a formula is used as below to approximate the number of electrons as\,\cite{smith2017development}:
\begin{equation}
    n_e = \frac{1}{h}I_{\text{Sun}}t_{int}^2\sigma_x\sigma_y\rho_x\rho_y\eta\times10^{-(m_{*}-m_{\text{Sun}})/2.5},
\end{equation}
where $h=$\SI{6.626070041e34}{\joule\second} is the Planck’s constant determining the relationship between energy and photons, $t_{int}$ is the integration/exposure time of the star tracker, $\rho_x = n_xx_p$ and $\rho_y = n_yy_p$ are the pixel sizes in two axes, and $\eta$ is an efficiency factor that accounts for quantum efficiency and lens efficiency. More explanations on the intermediate variables are shown in Fig\,\ref{fig:flux2eNumber}. Once the number of electrons is calculated, the area of pixels on the image plane taken up by a star can be determined.

\paragraph{Streak Simulation}
For optical-based space tracking, if RSOs move faster relative to the exposure time, the presence of RSOs in the optical images appears as streaks. These streaks can be modelled by straight lines convolved with the image PSF\,\cite{nir2018optimal}: 
\begin{gather}
    S = \varepsilon l(x_1, y_1, x_2, y_2) \circledast f(x,y),
\end{gather}
where $\varepsilon$ is the intensity of the streak in counts per unit length, $l$ is the single-pixel width line going from coordinates $(x_1, y_1)$ to $(x_2, y_2)$ , $f(x,y)$ is the PSF of the system expressed in Eq.\,\ref{eq:gaussianPSF} and $\circledast$ represents the convolution operator. Hence, with known pixel coordinates $(x_1, y_1)$ and $(x_1, y_2)$ at the starting and end epoch of the exposure duration, the streaks of the fast-moving RSOs can be simulated in the image. The stars, however, are still simulated as dots.

\subsection{Implementation}
The simulator is implemented via Matlab\textsuperscript{\textcopyright} classes, consisting of five main components:
\begin{itemize}
    \item \textit{clUSSIS}: representing the main class;
    \item \textit{clCatalogue}: obtaining astrometric positions and visual magnitudes of stars and RSOs;
    \item \textit{clSensorModel}: modelling optical sensors, e.g., ground-based telescopes and spaceborne star trackers;
    \item \textit{clNode}: calculating time-variant positions of observers (ground station or sensor-host satellite) and targets (stars and space objects);
    \item \textit{clPropagator}: inherited from the \textit{clNode} class to propagate orbits of the sensor-host satellite and space targets based on the Simplified General Perturbations (SGP4) propagator\,\cite{vallado2008sgp4} with TLEs.
\end{itemize}
Note that although the space image simulator is designed mainly to simulate star tracker images, it is also capable of simulating ground-based tracking images given the optical sensor properties. 

Four types of settings/parameters are loaded into the simulator to generate a synthetic image, including optics, detector, sensor and scenario settings. The boresight of the camera can be fixed in the inertial system or associated with the host satellite's attitude. 

The inputs for the star tracker simulator include:
\begin{itemize}
    \item target catalogue inputs: star catalogue with astrometric positions and visual magnitudes and RSO catalogue with orbital elements and radar cross-section areas;
    \item sensor model inputs: star magnitude limit;
    \item optics model inputs: number of length/width pixels ($n_{x}, n_{y}$), size of each length/width pixel ($x_p, y_p$) and focal length ($f$);
    \item image generation inputs: pixel spread, exposure time $t_{int}$, lens efficiency and quantum efficiency;
    \item boresight inputs: star tracker Euler angles $(\alpha_0, \delta_0, \phi_0)$.
\end{itemize}
The output of the space image is the list of stars within the FOV with the prescribed position and magnitude.

\subsection{Validation}
To validate the efficacy of star simulation, one image taken from the ASTERIA Cubesat\,\footnote{https://www.jpl.nasa.gov/missions/arcsecond-space-telescope-enabling-research-in-astrophysics-asteria} is used as a benchmark, following a similar process presented in Fig.\,4 of Ref.\,\cite{victor2021universal}. The date is chosen as the first day of 2018.
A snapshot of the sky taken by ASTERIA is given in the left subfigure of Fig.\,\ref{fig:ussisvalidation}, cropped from the spacecraft’s full FOV. 
The constellation Orion is in the bottom left. As a comparison, an image simulated is shown on the right of Fig.\,\ref{fig:ussisvalidation}. The resolution is $874\times738$ pixels with each pixel size of $2.8 \times 10^{-6}m$.
The vertical FOV equals to \SI{9.5}{\degree}. The magnitude threshold of 7 is used to limit the number of stars in the image. The boresight angles of the focal plane are listed as below:
\begin{equation}
    \begin{bmatrix}
        \alpha_0 = \SI{0.221}{\degree}, & \beta_0 = -\SI{3.667}{\degree}, & \phi_0 = \SI{0}{\degree}
    \end{bmatrix}.\notag
\end{equation}
The simulated image shows alignments of the stars with respect to the ASTERIA image (indicated by red ellipses), which manifests the efficacy of the simulator in star simulation.

\section{Evaluating DCNN-based Methods for RSO Streak Detection}\label{sec:framework}
After simulation, we are then able to perform more meaningful evaluation and comparison for RSO streak detection with different detection methods. In the following sections, we will briefly introduce the different detection methods we included for comparison and then describe the protocols we used for evaluation. 

\subsection{Overview of Detection Methods}\label{sect:DCNNover}
To detect RSO streaks from the space images, it is applicable to apply popular object detection methods for this task. Current mainstream object detection methods \cite{redmon2016you, he2016deep} are mainly based on DCNNs. In general, DCNNs are the models that can learn and identify complicated and high-level visual patterns in images by simulating the neural networks of human visual perception system. DCNNs usually have a deep structure (e.g., with a depth of more than 50 layers) and can achieve robust visual recognition performance on par with human beings. Using DCNNs, cutting-edge object detectors can be described by a two-phase procedure: feature extraction and object detection. For feature extraction, DCNNs translate input image data into feature maps at different smaller scales, like 4, 8, 16, 32 times smaller than the input image. Reducing the scales of feature maps is helpful as it saves great computational costs and extracts high-level and more robust feature representations. The DCNN used for feature extraction is usually termed as a "backbone". Then, with the extracted feature maps, a small neural network will scan these feature maps and fulfil the object detection task by estimating whether an object appears at a scanned location and how large this object is. Feature maps of different scales are usually used for detecting objects of corresponding sizes. It is worth mentioning that modern object detection algorithms diverge and follow different strategies to carry out the final detection. There are two major types: the first one tends to integrate the DCNNs for feature extraction and object detection into a unified end-to-end structure, i.e., the "single-stage" detector.
On the other hand, the second type of detector first roughly estimate the object locations (termed as "proposals") using the backbone and the neural network for scanning and then introduce an extra DCNN to refine the proposals. This type is usually called a "two-stage" detector. In practice, the single-stage detectors are usually more efficient but less robust, and the two-stage detectors are more robust but also more time-consuming.

\begin{figure*}[!t]
    \centering
    \includegraphics[width=\textwidth]{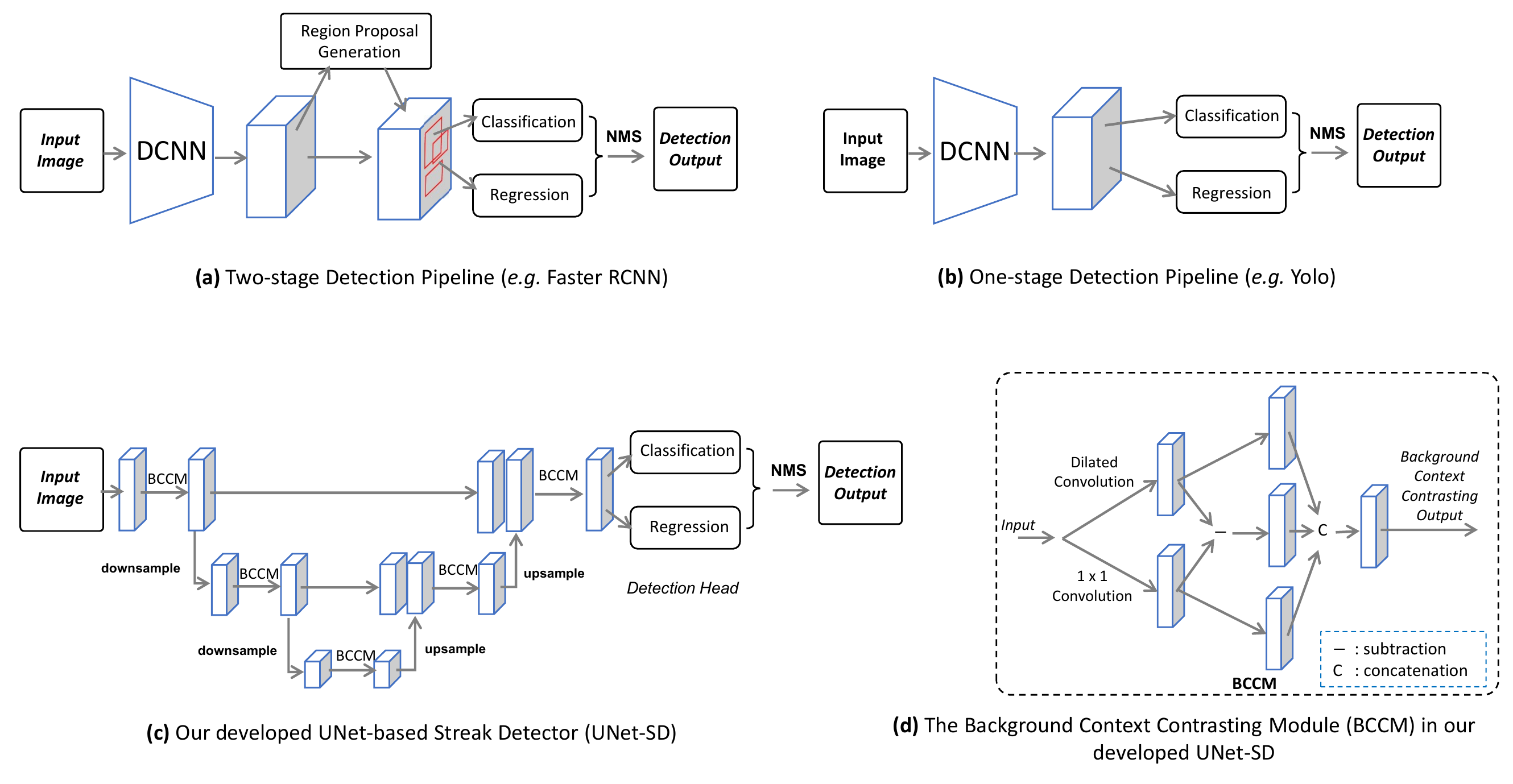}
    \caption{Conceptualized network structures that are evaluated under our SAB-RSOD in this study, including (a) two-stage detector like the Faster RCNN, (b) one-stage detector like the Yolo, and (c) our-developed UNet-based streak detector (UNet-SD). (d) shows the background context contrasting module (BCCM) devised to improve UNet-SD. }
    \label{fig:nets}
\end{figure*}

However, detecting RSOs, like satellites, from space images can be particularly difficult. Taking Fig.\,\ref{fig:exSwarmImg} as an example, it can be observed that the target RSO appeared as a streak dimmer than most background stars. In addition, the length of the streaks in different images varies significantly. As a result, it is challenging to detect the RSO robustly among all datasets, even with naked eye. In practice, DCNN-based detectors can perform as robustly as humans only if abundant data is available for training and the image resolution is sufficiently high. Fortunately, with our proposed simulation-augmented benchmarking framework as described earlier, we can gain access to rich simulated data that is similar to the real-world data for training detection methods sufficiently. Meanwhile, to tackle the problem of image resolution, we further developed a novel detection method, called UNet-SD, to perform high-quality detection if using low image resolution. The details about our developed UNet-SD can be found in Sec. \ref{sec:unet}. Fig. \ref{fig:nets} shows an overall pipeline comparison between the included detection methods. 

\subsection{Details about Evaluated Detection Methods}
In this study, we first introduce the two most popular object detectors to fulfil the RSO streak detection task, Faster RCNN \cite{ren2015faster}, and Yolo \cite{redmon2016you}. The Faster RCNN is the foundation of cutting-edge modern object detectors \cite{chen2019hybrid}, and the Yolo is the most famous efficient object detection method that has many derivative methods \cite{redmon2018yolov3,bochkovskiy2020yolov4} and has been widely applied in the industry. Although Faster RCNN and Yolo are popular and effective, they suffer from low-resolution and extremely small object sizes when using Swarm images. To tackle this, we further develop a UNet-based Streak Detection (UNet-SD) method which is able to detect small objects on low-resolution images effectively. 

\subsubsection{Faster RCNN}
The Faster Region-based Convolutional Neural Network (RCNN)\,\cite{ren2015faster} is taken as the most representative two-stage detection method. It first uses a DCNN to extract features into different sub-sampled levels. Then, it uses a region proposal network (RPN) to learn and generate proposal windows that may contain an object on the image. Afterwards, it applies a small Convolutional Neural Network (CNN) to classify the categories of the objects inside each proposal and regress the locations of the appeared objects inside. This forms a 'two-stage' detection pipeline.
Later, to tackle the problem that the vanilla Faster RCNN performs poorly on small objects, Feature Pyramid Network (FPN)\,\cite{lin2017feature} is introduced. The FPN makes use of all the feature levels extracted by the DCNN, so that higher-resolution features can be considered to detect small objects. Later, the integration of Faster CNN and FPN has been the basic component of state-of-the-art detectors\,\cite{chen2019hybrid}. For more details, please refer to the \cite{ren2015faster, lin2017feature}.

\subsubsection{Yolov3} 
Yolo, which is the abbreviation for You Only Look Once \cite{redmon2016you}, is one of the most famous efficient object detectors. It significantly simplifies the pipeline of Faster RCNN by directly predicting the locations and categories of appeared objects in the images based on the features extracted by a lightweight backbone DCNN. Compared to the two-stage detector, Yolo formulates a 'single-stage' detector. Due to its efficiency, the Yolo detector is widely used in many industrial applications. Since the introduction of the original Yolo object detector, there have been a number of improved versions released. \cite{redmon2018yolov3,bochkovskiy2020yolov4}. In this study, we investigate the more popular Yolo v3 detector on the RSO streak detection task. For more details, please refer to \cite{redmon2018yolov3}.

\subsubsection{Our developed UNet-based Streak Detector}
\label{sec:unet}
Using Yolo and Faster RCNN for RSO streak detection, we found that these popular detectors would meet significant challenges when using the Swarm data. More specifically, space objects usually have extremely small sizes with respect to the whole image size and are very dim in brightness. For example, a streak can only have 10 pixels in terms of height and width, while the whole image has a size of 576 by 768 on a Swarm image. Since the backbone design used in current detectors heavily rely on the down-sampled feature maps for detection, it is significantly difficult to depict extremely small objects on these feature maps. Although enlarging the input images can also enlarge object sizes, this will introduce a great extra computational complexity as well, making current detectors less practical for applying to embedding systems employed in the space. Moreover, the space objects can be very dim in the image. The brightness of some streaks can be too low to distinguish them from the background space areas. This issue can further confuse the current detectors between the streaks and constellations on the down-sampled feature maps. 
To tackle these problems, we attempt to develop a specifically designed detection method called UNet-based Streak Detector (UNet-SD).

In general, our UNet-SD is based on UNet \cite{ronneberger2015u}, which is a special DCNN structure that upsamples the down-sampled feature maps in DCNNs back to the original image size. Using subsequent down-sampling and up-sampling operations, the DCNN turns out to have a "U"-character shape, thus termed a "U" Net. The UNet allows us to detect objects at their original sizes on the images without down-sampling. However, only using the UNet is challenging to achieve robust streak detection due to the following reasons. Firstly, the UNet structure can be used for feature extraction, but it cannot output detection results directly. Secondly, it has a lower model capacity than DCNNs used in the Faster RCNN and the Yolo, which means that it is more difficult to achieve effective detection using the UNet. 

We design a new detection head to make it compatible with the UNet structure. The new detection head is defined by a 1 by 1 convolution which takes as input the UNet features and outputs two fields, i.e., a classification output to estimate the existence of space objects and a regression output to predict the sizes of an appeared space object. We make the network directly output the actual size of an object. This is fulfilled by normalizing the sizes of streaks \textit{w.r.t.} the image sizes. The normalization procedure is similar to the method \cite{liu2016ssd}, in which we divide the width and height of a bounding box by the width and height of the whole image.

In addition, to improve the model capacity of the UNet and improve the visual modelling on dim streaks, we introduce a background context contrasting module (BCCM) in each convolution layer in the UNet. The detailed design of the BCCM can be found in Fig. \ref{fig:nets}. In particular, we extract the local feature and its surrounding context features using 1 by 1 convolution and dilated convolution, respectively. Then, we extract the contrasting value by subtracting the local feature values from the context feature values. Later, we collect all these local, context, and contrasting features to form a more effective feature representation to refine the UNet features.

\subsection{Further Discussions}
Several solutions to the problem of streak detection have been suggested, in four general classes: simple source detection, computer vision code, machine learning algorithms, and template fitting to line shapes.

The literature review indicated a need for machine learning for RSO streak detection. However, to validate the need for machine learning initial investigation was complete using a traditional computer vision technique based on OpenCV library\,\cite{barodi2020applying} as well as the implementation of ASTRiDE, which is based on source detection\,\cite{kim2016astride}. The outcome of the ASTRiDE results is presented below for comparison purposes. 

Traditional computer vision techniques such as line detection can be used for streak detection; however, these techniques are not optimal in the context of RSO detection. For example, using functions in the OpenCV library, it is possible to detect the edges in an image and then feed this information into another function that uses probabilistic Hough line transform to detect lines. This technique relies on an obvious contrast between the RSO streak requiring detection and the image background. In the presented dataset, the faint nature of the streaks meant that many RSO would be missed and false positives would be given with other brighter pixels such as stars. This is an undesired outcome and requires further human intervention to further categorise the streaks. 

Besides the DCNN-based detectors outlined in section \ref{sect:DCNNover}, we also consider the streak detector based on traditional image processing techniques that do not involve any DCNN in their models. In particular, we apply the ASTRiDE for RSO detection. The ASTRiDE attempts to detect streaks in astronomical images based on the "border" of each object. They trace the long edges of the shape of each border to determine whether the border belongs to a streak. 

In practice, although ASTRiDE is effective for detecting artificial satellites from common astronautical images that have clear object boundaries, it performs poorly on Swarm images taken from a ground-based telescope. We suppose that this is because both data domains are quite different. The Swarm images tend to be contaminated by lower signal-to-noise ratios. As a result, the RSO detection for Swarm real-world images by ASTRiDE presents unsatisfactory results, including splitting an individual streak, picking up a wrong region for a streak, no output for streak centroid coordinate, etc. An example of a wrong detection by ASTRiDE for Fig.\,\ref{fig:exSwarmImg} is shown as below in Fig.\,\ref{fig:astrideRes}. The streak appears in the top middle of the figure. However, ASTRiDE detects another region in the bottom middle indicated by a blue circle and the number of $1$. 
When evaluating ASTRiDE, it can hardly provide meaningful results, while DCNN-based detectors can learn to detect small and dim streaks more easily and effectively. We thus mainly focus on studying the effects of different DCNN-based detectors regarding this task. This technique behaves well for detecting linear streaks in images with a sparse background. However, it is not well suited to images that appear to have cluttered backgrounds with faint streaks, as demonstrated in Fig.\,\ref{fig:astrideRes}. Hence the rationale for using machine learning techniques for detecting streaks in a set of images with variations in the environment.
\begin{figure}[!ht]
    \centering
    \includegraphics[width=0.4\textwidth]{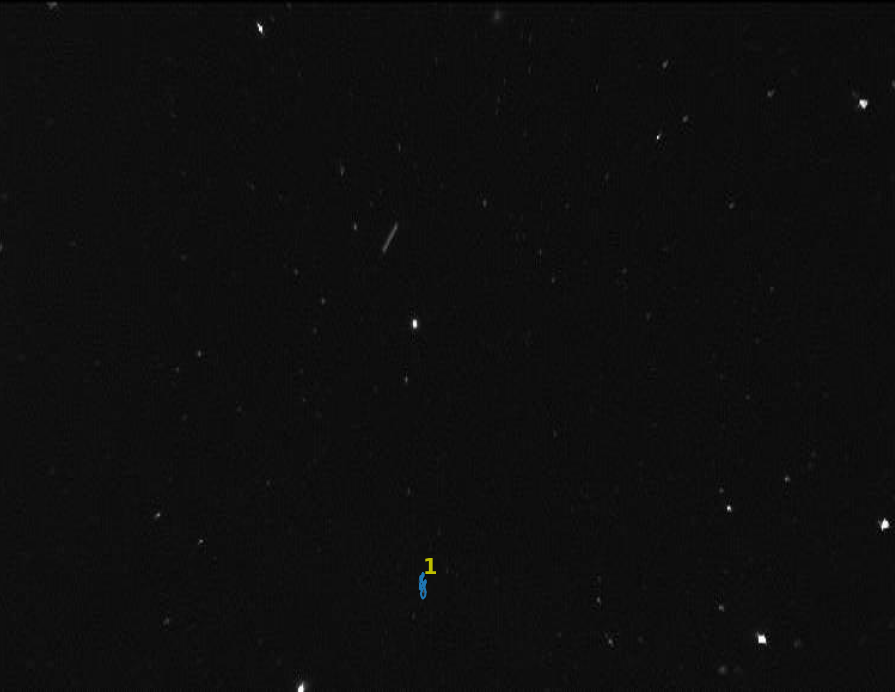}
    \caption{One Exemplar Result by ASTRiDE}
    \label{fig:astrideRes}
\end{figure}

\subsection{Evaluation Protocol} 
In this section, we will discuss in detail how we train and test the aforementioned detection methods under our proposed benchmarking framework. In particular, we will first introduce the data processing and training strategies we used for evaluation. Then, we describe the evaluation metrics we designed to better reveal the performance on the RSO streak detection task. 

\subsubsection{Data Processing and Training Strategy}
To test different methods for RSO detection effectively, we mainly use both real-world and synthetic data for training. Given the fact that large volumes of real-world images taken by space-borne sensors are hardly accessible, synthetic images generated by the software can provide a good complement to help tackle the lack of data issue and better train DCNN-based detection models. To achieve this, we develop a specific data wrapping module that standardizes the image and annotations of the real-world Swarm data, as well as the synthetic data. The data wrapping module also mixes both the Swarm and synthetic images for training. All the considered images will also be resized to a unified size for both training and testing. 
To train different DCNN-based detection methods well and fairly, we adopt several data augmentation methods during the training phase. These data augmentation methods include image flipping, multi-scale resizing, cropping, shifting, color space disturbing, brightness disturbing, and so on. For example, take an image as shown in Fig.\,\ref{fig:exSwarmImg}, we can randomly crop a smaller patch from this image to form a new image which is then resized to the unified size and fed to the detection model for learning. Note that we apply the same data augmentation methods to different evaluated detectors for a fair comparison. The random seed is also fixed for training different models.

To optimize different DCNN-based detectors properly, we use the stochastic gradient descent (SGD) algorithm with a learning rate of 1e-5. We train each evaluated detector with 10k iterations and drop the learning rate by a factor of 0.1 at 5k-th and 9k-th iteration. We empirically choose a weight decay parameter of 5e-4 to help prevent overfitting. These training parameters will also be fixed when training different detectors. Regarding the loss functions that define the desired outputs of DCNN-based detectors for optimization, we follow the default ones as described in the related studies\cite{ren2015faster}. For the developed UNet-SD method, we follow the loss functions of Yolo series \cite{redmon2016you,redmon2018yolov3}, except that it uses a longer training period. This is because we use the model parameters pre-trained on ImageNet\,\cite{deng2009imagenet} for training Faster RCNN and Yolov3 detectors, which can directly accelerate their convergence speed. On the contrary, the implemented UNet-SD is trained from scratch, thus we increase its training period to 40k iterations to ensure proper convergence for a fairer comparison.

During the experiment, we use the server that contains Tesla V100 GPUs for training and testing. This hardware environment is kept the same for all the evaluated algorithms. 

\subsubsection{Evaluation Metrics}
To evaluate the RSO detection performance, we mainly rely on the Swarm data because we would like to know the actual performance of evaluated detectors on real-world images. Besides, we found that there could still be a considerable gap (or visual differences) between the Swarm data and the simulated data, which makes the evaluation of the simulated data less meaningful to reveal the performance of the real-world data. Then, to evaluate the limited Swarm data, we attempt to perform a 5-fold cross-validation process. The 5-fold cross-validation process is a commonly used cross-validation technique to test model performance in statistics and artificial intelligence. It randomly and evenly splits a dataset into 5 different groups and it involves 5 rounds of training and evaluation procedure. In each round, one group is selected as the test dataset and the other 4 groups are considered as the training dataset. Lastly, the evaluation performance obtained from all 5 rounds are averaged to obtain the final results. Through 5 rounds of training and evaluation, every group will have a chance to become the test dataset, making the estimate of the mean model performance more representative.

During the evaluation, we mainly follow the tradition of object detection studies\,\cite{ren2015faster,redmon2016you} and use Average Precision (AP) as the main metric to reveal model performance. To compute AP, we would like to first introduce the concept of Intersect-over-Union (IoU) in object detection. 
The IoU is computed based on the proportion of overlap areas \textit{w.r.t.} all the areas covered by two bounding boxes. It reveals how close the two bounding boxes are. Therefore, with the annotated desired detection outputs (termed as ground-truths), we can estimate the quality of detection results \textit{w.r.t.} the provided ground-truths. 
In particular, if a detection result has a high IoU \textit{w.r.t.} a ground-truth, this detection result can be considered a success, and \textit{vice versa}. For example, an IoU threshold for determining whether a detection result is successful can be set as 0.5. By identifying successful detection and failed detection, we can obtain the number of true positive (TP) and false positive (FP) detection results. With these statistics, we can then compute a precision score according to:
\begin{equation}
    Precision = \frac{TP}{TP+FP}.
\end{equation}
With this precision calculated for each image, we can then evaluate the overall performance of a detection algorithm using by calculating the mean precision of all the considered images. Normally, the AP at a threshold of 0.5 (denoted as "AP@0.5") can already reveal the detection performance effectively. Besides, since the streaks are usually very small and are difficult to detect precisely, we also use the AP at 0.3 (denoted as "AP@0.3") to estimate whether a detector can at least find out the rough area of streaks. 
Meanwhile, the IoU threshold can also range from 0.5 to 0.95 to evaluate the overall detection quality. The threshold of 0.95 means that only detection results that strictly align with a ground-truth box can be considered a successful detection. By averaging the precision at different thresholds, we can obtain the overall AP (denoted as "AP@[0.1:0.95]") which can comprehensively reflect the effectiveness and robustness of a detection algorithm. In practice, these AP metrics are widely applied in current detection applications and research. For more details, we refer readers to \cite{lin2014microsoft, everingham2010pascal}. 

In addition to the AP, we also attempt to study the efficiency of the compared object detection algorithms. In general, there are several factors that can indicate how efficient an object detector is, including GFLOPs (Giga Floating Point Operations) and inference time. It indicates the amount of floating point operations used in an artificial intelligence algorithm and thus can be used to estimate the computational complexity of a model. We also study the inference time of the compared algorithms. The inference time refers to the actual time cost of an algorithm. This directly relates to the computational speed of an algorithm. Although inference time can be different from GFLOPs due to diversified hardware environments, both the GFLOPs and inference time are correlated. Meanwhile, in order to study the speed/accuracy trade-off of different object detection algorithms, we attempt to vary the image sizes during evaluation. More specifically, we rescale the images used for training and testing into different sizes in different experiments. Usually, larger images mean higher computational complexities and costs, as well as better accuracy due to more pixel details. By enlarging images, we can estimate the improvement in the accuracy of an algorithm, which can reveal its true potential for robust detection. Lastly, we also study the number of parameters in different algorithms.

\section{Experiments and Results}\label{sec:exp&res}
In this section, we present the evaluated results of different considered detection methods on the RSO streak detection tasks under our proposed benchmarking framework. More specifically, we first present the statistics of synthetic images compared to the Swarm dataset. Then, we compare the detailed detection performance of evaluated detection methods and their detection efficiency. Later, we further illustrate the effects of other factors like different image resolutions, whether using the simulated data for training and how many simulated images are used for training has an effect. Lastly, we discuss some qualitative results of detection.


\subsection{Statistics of Simulated Images and Swarm Dataset}
\begin{figure}[!ht]
    \centering
    \begin{subfigure}[b]{0.497\textwidth}
        \centering
        \includegraphics[width=\textwidth]{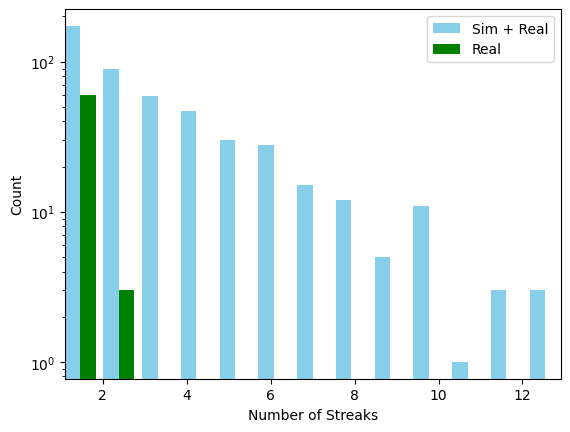}
        \caption{Comparison of Numbers of Streaks in Each Image between Synthetic and Swarm Star Tracker Images.}
        \label{fig:cmpNumStreaks}
    \end{subfigure}
    \hfill
    \begin{subfigure}[b]{0.497\textwidth}
        \centering
        \includegraphics[width=\textwidth]{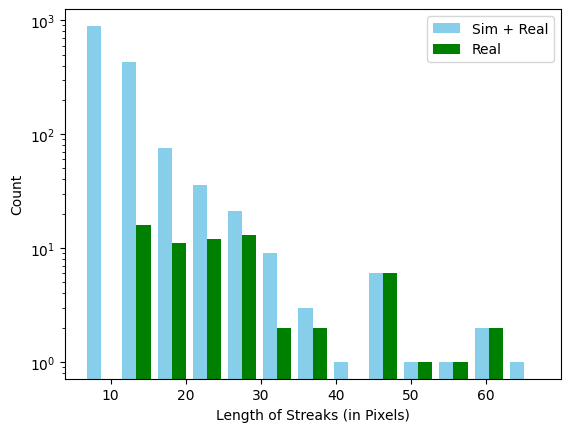}
        \caption{Comparison of Streak Lengths between Synthetic and Swarm Star Tracker Images.}
        \label{fig:cmpLenStreaks}
    \end{subfigure}
    \caption{Statistics of Simulated Images and Swarm Dataset}
    \label{fig:imgStatistics}
\end{figure}

Based on the simulation, we have included 400 synthetic images that contain clear simulated streaks besides the 63 real-world Swarm star tracker images. The number of streaks in each image and the pixel length of each streak are compared between simulated plus real-world images with real-world ones only, shown in Fig.\,\ref{fig:cmpNumStreaks} and Fig.\,\ref{fig:cmpLenStreaks}, respectively. It is seen that the diversity on streaks has been improved via simulation, especially more shorter streaks have been generated into the training pool. This greatly alleviates the problem of data shortage for training DCNN models, which is verified by the experimental results discussed later. 

\subsection{Detection Results and Benchmarking}\label{sect:redults}
Using the simulation-extended dataset, we evaluate different detection methods using AP@[0.3:0.95], AP@0.3, AP@0.5. In this section, we present the experiments on the overall comparison, speed-accuracy trade-off, ablation studies, and qualitative studies. In these experiments, we thoroughly investigate the difference in the detection performance of the included methods, i.e., Yolo, Faster RCNN, and UNet-SD, in different settings, so that the advantages and disadvantages of different methods can be revealed. We also explore the effects of simulation in the benchmarking process to demonstrate that simulation-based dataset augmentation is important for RSO streak detection. 

\subsubsection{Overall Benchmark Results}
We first present the overall benchmarking output of the compared detection methods. Table\,\ref{tab:ovrall} shows the statistics. In this experiment, we use the input image with a resolution of 576 by 768. All detectors follow the same training strategies as described previously, and all the simulated images are included for training. In the table, we report all the considered metrics, including AP@[0.3:0.95], AP@0.3, AP@0.5, GFLOPs, parameter numbers, and inference time. The reported performance is averaged across 5-fold cross-validation.

From the results, we can obtain the following findings. Firstly, all the methods have a good performance for AP@0.3, which means that all the compared methods perform well under loose detection criteria. If we make the criteria stricter, e.g., using AP@0.5, the performance of Yolo and UNet-SD drops drastically by around 0.2 and 0.1 in AP, respectively. This suggests that the two detectors can identify the streaks properly but the detected bounding boxes may not be always accurate. Instead, the Faster RCNN performs well for both AP@0.3 and AP@0.5, showing that this detector is more consistent than the other two detectors and is less likely to produce inaccurate detection boxes. This is easy to understand because the Faster RCNN has a specific classification head that can clearly identify partially aligned less accurate bounding boxes. Regarding the overall performance for AP@[0.3:0.95], the Faster RCNN shows much better consistency than the other two detectors, which further validates its superior robustness. In addition to the accuracy, if considering complexity like GFLOPs, parameters, and inference time, it shows that the Faster RCNN is the heaviest method with the highest GFLOPs, the highest number of parameters, and the longest processing time, while the UNet-SD is the most light-weighted. Yolo is faster than Faster RCNN, but its improved modules introduced in \cite{redmon2018yolov3} bring more parameters. 

To sum up, for the RSO streak detection task, our benchmarking results illustrate that the Faster RCNN is the most robust and accurate method but it has a higher cost. Our developed UNet-SD method is very efficient and also produces promising performance if not pursuing the highest accuracy. Yolo is roughly in the middle of the two methods, but its accuracy in the standard of AP@0.5 is greatly affected by the low-resolution of input images and small sizes of streaks.

\begin{table*}[!t]
    \caption{Overall results of compared algorithms in terms of Average 
    Precision (AP), GFLOPs, parameter number (Params), and inference time (Time). }
\resizebox{\textwidth}{!}{%
    \begin{tabular}{l|c|c|c|c|c|c}
    \hline
        ~ & AP@[0.3:0.95] & AP@0.3 & AP@0.5 & GFLOPs & Params (millions) & Time  (ms) \\ \hline
        Yolo \cite{redmon2016you}   & 0.39 & 0.78 & 0.59 &84 & 1.3 & 25  \\ \hline
        Faster RCNN \cite{ren2015faster, lin2017feature}  & 0.85 & 0.86 & 0.86 & 97& 61.9 & 31  \\ \hline
        UNet-SD (our developed) & 0.52 & 0.93 & 0.85 & 10 & 41.5 & 20  \\ \hline
    \end{tabular}
    \label{tab:ovrall}
}
\end{table*}

\subsubsection{Amounts of Simulation Images}
\begin{figure*}[!th]
\centering
\includegraphics[width=\textwidth]{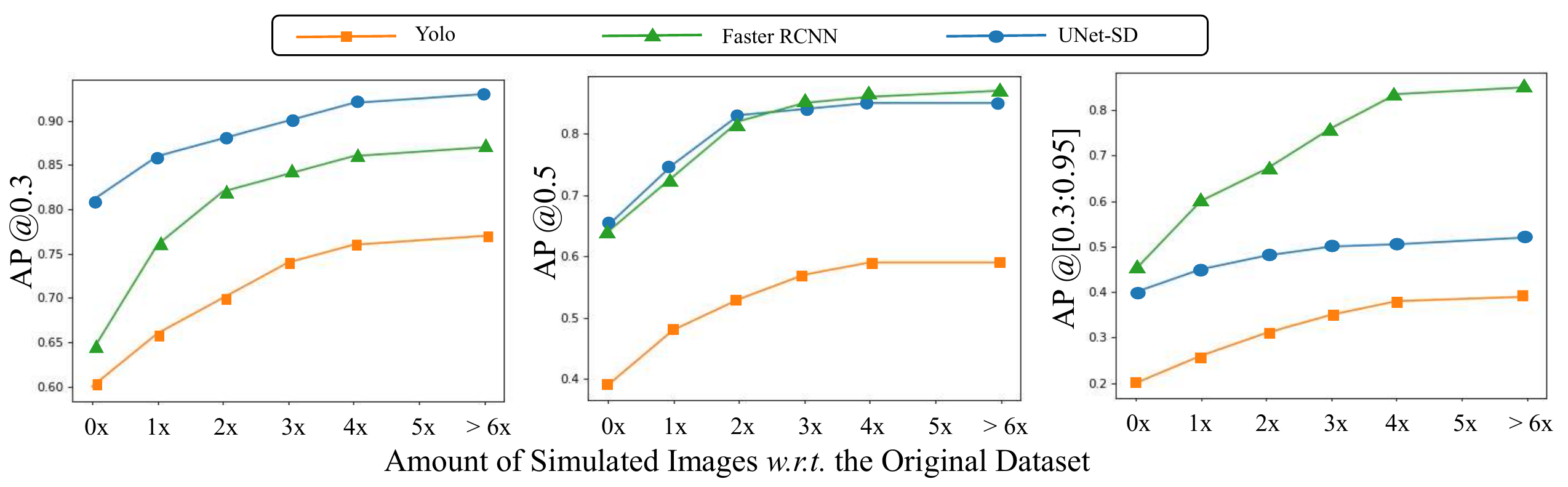}
\caption{Speed-accuracy tradeoff study. The "s1", "s2", "s3" respectively means using the input image with a resolution of 576 x 768, 608 x 800, and 640 x 864. Larger resolution usually means more computational costs and better accuracy.}
\label{fig:amount}
\end{figure*}

In our proposed benchmarking framework, the simulated images play an important role in effective training on different detectors. Here, we study the effects of different amounts of simulated images included to augment the original dataset. Figure\,\ref{fig:amount} shows the detailed results. In addition to the original dataset which only has around 50 real-world images for training and 10 real-world images for testing, we attempted to add 400 simulated images to augment the original dataset for training. 
All performances are 5-fold cross-validation results. In this experiment, we still report all the AP@0.3, AP@0.5, and AP@[0.3:0.95] results. 

From the figure, we can observe that increasing the amount of used simulated images can all lead to improved performance, which demonstrates that the simulation is effective for enriching the original dataset with a very limited amount of training images. It is worth noting that the Faster RCNN improves the most with the help of simulated images when using AP@0.5 and AP@[0.3:0.95]. Our developed UNet-SD can perform well with limited images, but it quickly saturates when increasing the amount of simulated images considered for training. For the Yolo, even using additional simulated images for training, it still struggles to achieve high-quality detection.

\subsubsection{Speed-Accuracy Trade-off Comparison}
\begin{figure*}[!th]
\centering
\includegraphics[width=\textwidth]{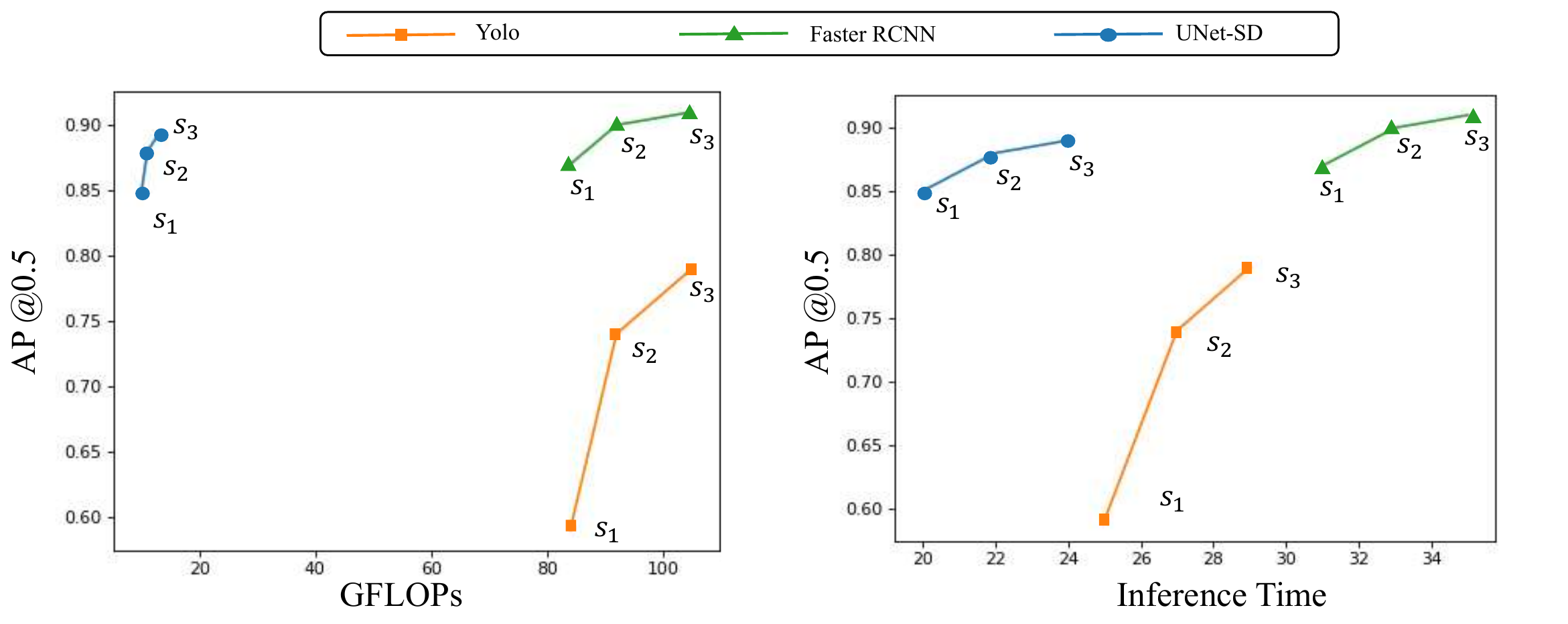}
\caption{Speed-accuracy tradeoff study. The "s1", "s2", "s3" respectively means using the input image with a resolution of 576 x 768, 608 x 800, and 640 x 864. Larger resolution usually means more computational costs and better accuracy.}
\label{fig:spacc}
\end{figure*}

Here, we also study the speed-accuracy tradeoff and the results are shown in Fig.\,\ref{fig:spacc}. We have studied the model performance using the input image with a size of 576 by 768, 608 by 800, and 640 by 864, respectively. The original Swarm data has a size of 576 by 768 and we adopt the wide-used bilinear interpolation to obtain images with higher resolutions. 
Moreover, we use both the GFLOPs and inference time to represent the speed, so that a more comprehensive comparison can be achieved. 

From the results illustrated in Fig.\,\ref{fig:spacc}, we can observe that increasing the image size or resolution can indeed improve performance promisingly. In particular, all the compared algorithms achieve the highest AP scores using the largest input image. For a detailed detection algorithm comparison, we can conclude the following points based on the obtained results. Firstly, our developed UNet-SD is the most efficient algorithm among compared algorithms, while the Faster RCNN is the slowest. However, when increasing the image resolution, Faster RCNN achieves the highest accuracy. Furthermore, the Yolo also improves significantly for higher-resolution images. We analyse that this can be attributed to two main reasons: 1) increasing the image resolution also enlarges space objects, thus the down-sampled feature maps in current popular detectors like Yolo and Faster RCNN can represent the enlarged space objects more properly, making it much easier to detect space objects effectively; 2) increasing the image resolution can introduce more pixel details and the expressive capacity of the backbones of popular detectors can be maximized to model the visual patterns with more details. It is worth mentioning that the backbones of Yolo (i.e., darknet\,\cite{redmon2016you}) and Faster RCNN (i.e., ResNet\,\cite{he2016deep}) usually have around 50 processing layers, while the UNet-SD only has around a dozen processing layers. As a result, the benefit of a deeper backbone can be magnified by increasing image resolution. One of the negative effects of increasing the image size is that the computational costs would also increase greatly. This aligns with the findings proposed in the study\,\cite{huang2017speed}.

In general, increasing the image resolution can improve the accuracy of a detector due to more pixel details provided but can also introduce excessive computational complexity. 

\subsubsection{Extra Ablation Study on UNet-SD}
In our developed UNet-SD, we have introduced a background context contrasting module (BCCM) to improve the model capacity of the UNet. In the experiment, we have found that the BCCM is important for the UNet-based detector to help achieve high-quality RSO streak detection. In particular, without the BCCM, only using a pure UNet structure \,\cite{ronneberger2015u} and a single-stage detection head would achieve a performance of 0.32, 0.91, 0.70 in terms of AP@[0.3:0.95], AP@0.3, and AP@0.5, respectively. Note that our developed UNet-SD with the BCCM achieves a performance of 0.52, 0.93, 0.85. As a result, the BCCM is beneficial for improving the accuracy of the detection results obtained from UNet.

\subsubsection{Qualitative Analysis}

\begin{figure*}[!t]
\centering
\includegraphics[width=\textwidth]{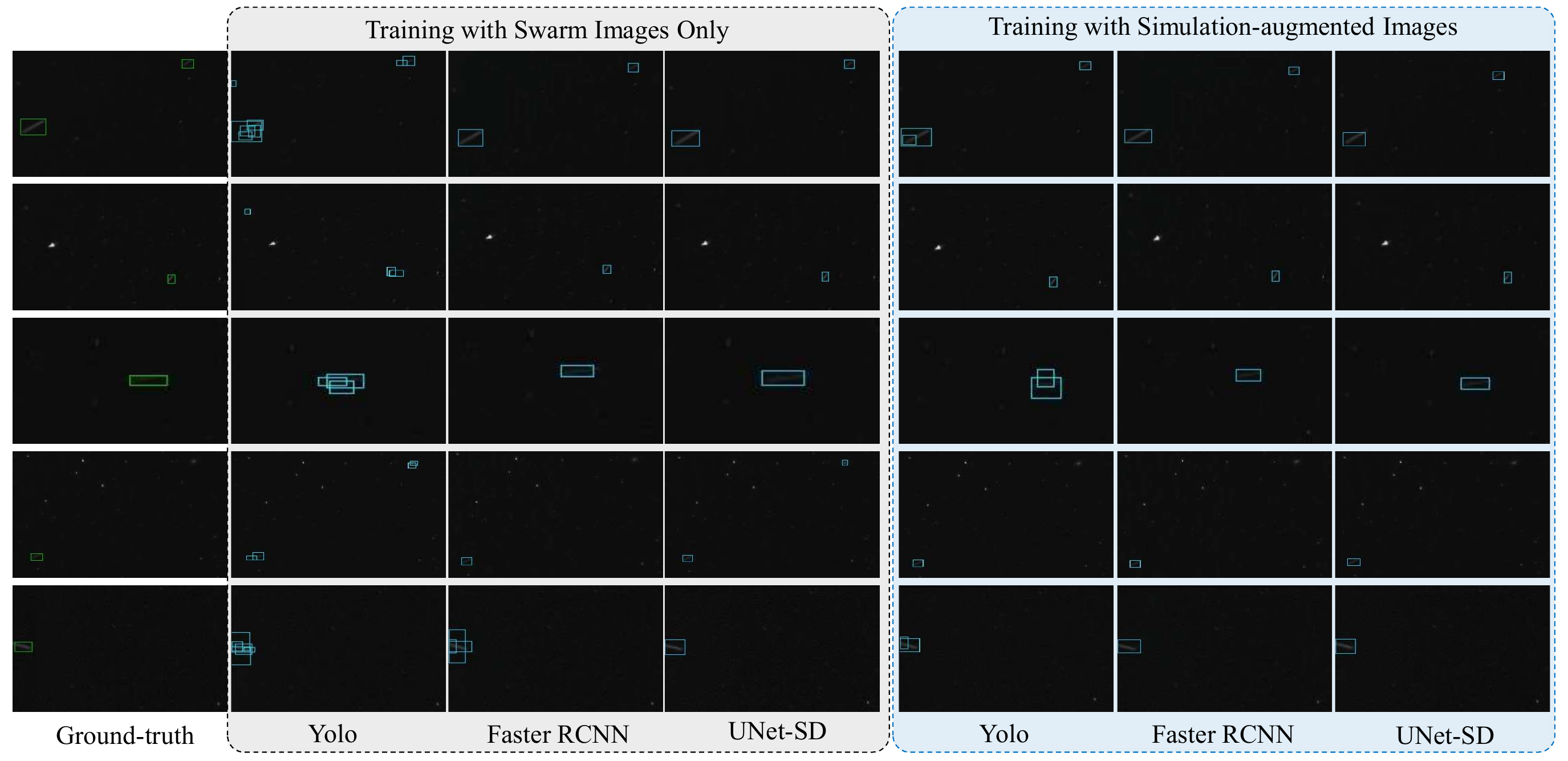}
\caption{Qualitative results of different compared algorithms together with the input image and ground-truths. All detection results are illustrated using light-colored rectangular bounding boxes.}
\label{fig:qua}
\end{figure*}

Lastly, we show the actual detection outputs and qualitative results of all the compared algorithms in Fig.\,\ref{fig:qua}.
In the figure, the input images and ground-truths are also presented as a reference. We draw the detection outputs of different algorithms using light-colored bounding boxes in the images. 4 images from the test set of a training and evaluation process are selected as examples and are used for illustration. 

From the presented figures, we can observe that the space objects, or streaks, are indeed extremely small and dim in the image. Using the traditional image processing technique, the ASTRiDE can hardly detect space objects properly. Instead, we can find that deep learning algorithms can all deliver meaningful detection results in general. For the Yolov3, it can identify all the space objects but the bounding boxes are quite inaccurate. This may because the appearances of space objects are significantly down-sampled in the backbone network, making the Yolov3 detector difficult to describe these space objects precisely. Using the Faster RCNN with FPN that introduces another stage for refinement, we can find that the detection results can more accurately cover the space objects. However, since the Faster RCNN with FPN also relies on down-sampled feature maps, this detector still struggles to describe space objects correctly. In particular, it generates two overlapped bounding boxes in the first row and misses the space objects in the second row. By using high-resolution UNet-based detector, more pixel details are considered for detection, making it more accurate to detect space objects. Despite the promising performance in the illustrated images, we would like to mention that the high-resolution UNet-based detector sometimes produced more false positive detection results during the experiment. We analyse that this may because the UNet backbone is too shallow and thus its expressive capacity is too limited to distinguish visual patterns within the false positive detection results. 

Based on the obtained results, we can first conclude that the current cutting-edge artificial intelligence and deep learning-based object detection algorithms are effective for RSO detection. In this report, we demonstrated that the Yolov3, Faster RCNN with FPN, and the high-resolution UNet-based detector are superior to the ASTRiDE that relies on traditional image processing techniques. In addition, we can also find that sufficient training data and powerful computational ability are two key factors for high-quality RSO detection. By simulating more data to facilitate training, all the deep learning-based detectors can deliver much more accurate detection results. Additionally, increasing the image resolution, which introduces plenty of extra computational costs, can also improve detection performance significantly. In addition to the data and computation, we also illustrate that re-designing a detection algorithm for the RSO detection task specifically is another direction to achieve more accurate detection. By recasting the current deep learning-based detection pipeline into the high-resolution UNet-based detector, we find that the small object detection problem can be alleviated and promising results can be obtained.



\section{Conclusion}\label{sec:conclusion}
This paper introduces a novel simulation-augmented benchmarking framework to detect resident space objects in space images based on deep convolutional neural networks. It is the first time the real-world Swarm star tracker images are used to train DCNNs augmented with images simulated by a high-fidelity simulator. 

For the feasibility of applying powerful deep learning-based object detection algorithms to the RSO detection task, we have the following findings. Firstly, it is important to exploit parallel computation using hardware like GPUs to make deep learning-based detection algorithms feasible for this task as using a CPU-only environment cannot satisfy the demand of real-time processing. Secondly, the speed-accuracy trade-off should be carefully considered when implementing a detection algorithm for the RSO detection task. Better detection is usually accompanied with higher computational costs, thus it is necessary to define how good a detector should be when estimating whether it is feasible for RSO detection. Lastly, if computational resources and speed-accuracy trade-off are well-considered. It is possible to develop novel detection algorithms that can be more efficient and feasible for this task. 
{The benefits of which outweigh the requirement for additional efforts on designing, research and testing. Therefore, providing a solution to object detection for SSA.}

The results indicate a strong probability of success when using UNet for RSO detection in space based applications. Future work will add the RSO detection framework developed by this work into the process of RSO astrometric positioning. Hence such serendipitous space-based optical observational data can be more beneficial to space situational awareness. The advent of mega-constellation, such as SpaceX's Starlink satellites, may provide potential future platforms for facilitating more space-based space tracking images. 


%


\section*{Acknowledgment}
The authors would like to thank SmartSat CRC for supporting this work. The background of this work was a joint project among The University of Sydney, Thales Australia and HEO Robotics. The input from the individuals on behalf of the industry partners is greatly appreciated. 

\section*{Appendix}
\label{sec:appendix}

\begin{figure}[!h]
    \centering
    \begin{subfigure}[b]{0.497\textwidth}
        \centering
        \includegraphics[width=\textwidth]{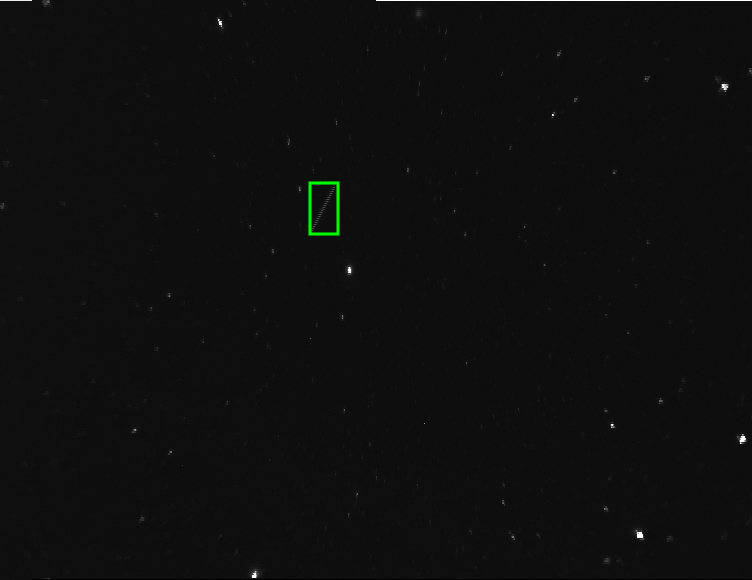}
        \caption{Interleaved image with RSO streak annotated by green box.}
        \label{fig:imgInterleaved}
    \end{subfigure}
    \hfill
    \begin{subfigure}[b]{0.497\textwidth}
        \centering
        \includegraphics[width=\textwidth]{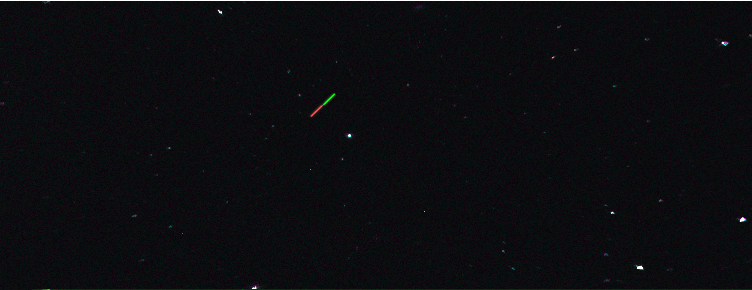}
        \caption{Difference of two single field images with RSO streaks indicated by red and green colours.}
        \label{fig:imgDifference}
    \end{subfigure}
    \caption{An exemplar image of real-world Swarm image with the green box annotating the space object as a streak.}
    \label{fig:exLongStreak}
\end{figure}
The Swarm dataset introduced in Sec.\,\ref{sec:statePro} comprises 2,504 images taken between 2015-2018. According to Feiteirinha et al., images from Swarm CHUs are made of two interleaved fields \cite{feiteirinha2019str4sd}.
Each one integrates sequentially for \SI{0.5}{\second} leading to a total integration period of \SI{1.0}{\second} for the entire image. Hence, three types of outputs are generated including interleaved images (see Fig.\,\ref{fig:imgInterleaved} for example), very short animations and difference images (see Fig.\,\ref{fig:imgDifference} for example). The streak in Fig.\,\ref{fig:imgInterleaved} represents an RSO and is annotated by a green box, which is less bright than many background stars. Streaks from two interleaved fields are distinguished by red and green colours in Fig.\,\ref{fig:imgDifference}, each of which is shorter than both together one. 
The entire image is preferable for RSO detection with a longer integration time, namely longer streaks, but the images from each individual field can also be included to add more variations to the dataset. 
After extracting single frames from the short animation, we collected 63 with at least one clear streak which was annotated manually. Note each animation file is normally composed of two frames. However, if one extracted frame does not present any streak it is discarded. 





\bibliographystyle{elsarticle-num-names} 
\bibliography{refs}


%
%
%

\end{document}